\documentclass[lettersize,journal]{IEEEtran}

\usepackage{amsmath,amsfonts,amssymb}
\usepackage[noend]{algorithmic}
\usepackage{algorithm}
\usepackage{array}
\usepackage[caption=false,font=normalsize,labelfont=sf,textfont=sf]{subfig}
\usepackage{textcomp}
\usepackage{stfloats}
\usepackage{url}
\usepackage{graphicx}
\usepackage{pdfpages}
\usepackage{cite}
\usepackage{bm}
\usepackage{booktabs}
\usepackage[table]{xcolor}
\usepackage{multirow}
\usepackage{mathtools}
\usepackage[above,below]{placeins}
\usepackage{tikz}
\usepackage{pgfplots}
\pgfplotsset{compat=1.16}
\usepgfplotslibrary{groupplots}
\usetikzlibrary{arrows.meta,positioning,shapes.geometric,fit,backgrounds,patterns,calc,decorations.pathreplacing}
\usepackage{tikz-3dplot}

\newcommand{\bR}{\mathbf{R}}
\newcommand{\bp}{\mathbf{p}}
\newcommand{\bv}{\mathbf{v}}
\newcommand{\bg}{\mathbf{g}}
\newcommand{\bn}{\mathbf{n}}
\newcommand{\bx}{\mathbf{x}}

\newcommand{\bH}{\mathbf{H}}
\newcommand{\bI}{\mathbf{I}}

\newcommand{\bP}{\mathbf{P}}
\newcommand{\bG}{\mathbf{G}}
\newcommand{\bLam}{\boldsymbol{\Lambda}}
\newcommand{\bSig}{\boldsymbol{\Sigma}}
\newcommand{\bb}{\mathbf{b}}

\newcommand{\diag}{\operatorname{diag}}
\newcommand{\tr}{\operatorname{tr}}

\newcommand{\SO}{\mathrm{SO}(3)}
\newcommand{\skewop}[1]{\lfloor #1 \rfloor_\times}
\newcommand{\Exp}{\mathrm{Exp}}
\newcommand{\Log}{\mathrm{Log}}

\newcommand{\bxhat}{\hat{\mathbf{x}}}
\newcommand{\Phat}{\hat{\mathbf{P}}}
\newcommand{\trigeq}{\triangleq}
\newcommand{\bplus}{\boxplus}
\newcommand{\bminus}{\boxminus}
\definecolor{bestbg}{RGB}{200,230,201}
\newcommand{\best}[1]{\cellcolor{bestbg}#1}
\definecolor{secondbg}{RGB}{255,236,179}
\newcommand{\second}[1]{\cellcolor{secondbg}#1}
\usepackage[colorlinks=true,linkcolor=blue,citecolor=blue,urlcolor=blue]{hyperref}

\begin{document}

\title{SA-LIVO: Efficient LiDAR-Inertial-Visual Odometry
       with Subspace-Aware Information Fusion}

\author{Yinong Cao, Xin He, Yijie Wang, Tianzhen Ma, Pingfeng He,
       Chenyang Zhang, Tingwei Wang, Shouzheng Zhu, Senyuan Wang, Changhui Jiang,
       Yuwei Chen, Chunlai Li, Genghua Huang, and Jianyu Wang%
\thanks{This work has been submitted to the IEEE for possible publication. Copyright may be transferred without notice, after which this version may no longer be accessible.}%
\thanks{%
(Corresponding authors: Xin He; Chunlai Li.)}%
\thanks{Yinong Cao, Yijie Wang, Tianzhen Ma, Chunlai Li, Genghua Huang, and
Jianyu Wang are with the Shanghai Institute of Technical Physics, Chinese
Academy of Sciences, Shanghai 200083, China (e-mail: lichunlai@mail.sitp.ac.cn).}%
\thanks{Xin He, Pingfeng He, Chenyang Zhang, Tingwei Wang, Shouzheng Zhu, Senyuan Wang,
Changhui Jiang, and Yuwei Chen are with the Hangzhou Institute for Advanced
Study, University of Chinese Academy of Sciences, Hangzhou 310024, China
(e-mail: xinhe@ucas.ac.cn).}}

\markboth{}%
{Cao \MakeLowercase{\textit{et al.}}: SA-LIVO: Efficient LiDAR-Inertial-Visual Odometry with Subspace-Aware Information Fusion}

\maketitle

\begin{abstract}
Light detection and ranging (LiDAR)-inertial-visual odometry fuses
geometric depth with visual measurements, but its sensors fail independently: LiDAR in under-constrained
geometry, vision under poor illumination or texture.
Weak information is harmless, since the filter then relies on the other
sensor or the inertial prediction; over-confident information is not.
Existing countermeasures act on weak information or switch vision by hand-set
thresholds.
This article presents SA-LIVO, which fuses vision through explicit,
online-estimated error models.
A frame-level noise scale, estimated by covariance matching, absorbs
whole-image errors, and Schmidt filtering carries the correlation of reused
observations.
Subspace-aware information fusion models the error shared by a frame's
observations as a pose offset capping the visual information per direction;
the exact fusion attenuates vision only above that cap and reduces to the
information sum when the visual model is consistent.
LiDAR and visual residuals share one invariant extended Kalman filter loop.
On 24 handheld and aerial sequences, SA-LIVO is the most accurate or tied on
22 and stays bounded where others diverge; on the HILTI 2022 benchmark it
needs 13.4~ms per frame on a laptop processor and $2.0$--$6.2\times$ less
memory than FAST-LIVO2, R3LIVE, and SR-LIVO.

\end{abstract}


\section{Introduction}
\label{sec:intro}

\IEEEPARstart{H}{andheld} scanners and small aerial vehicles are increasingly
used for surveying, inspection, and mapping where satellite positioning is
unavailable, such as building interiors, stairwells, and tunnels.
Their odometry supplies the pose that the planner and controller act on and
with which the map is built, and it runs on the platform's own computer for
the whole mission.
LiDAR-inertial-visual odometry (LIVO) tightly couples LiDAR depth, camera
appearance, and the motion prior of an inertial measurement unit
(IMU)~\cite{zheng2022fastlivo2,lin2022r3live,shan2021lvisam} and, in
structured environments, far exceeds single-modality accuracy.
Its exteroceptive sensors, however, degrade for independent reasons: the
LiDAR when the scene geometry leaves a pose direction unconstrained, the
camera when darkness, blur, or missing texture suppress image gradients.
Concurrent degradation is therefore common where these platforms operate,
and existing tightly coupled systems often diverge in it.
Two problems remain.

\emph{1) Over-confident visual information.}
In an information-form filter, a weak direction is carried by the other
sensor or the inertial prediction, so weak information alone is harmless.
Information that is wrong but confident is not: it dominates every
direction it constrains.
An error shared by all observations of a camera frame does not average
out, so the nominal visual information grows with every tracked point while
the information about the pose saturates.
Existing countermeasures do not model this error: degeneracy detection
rejects or constrains an ill-conditioned LiDAR
update~\cite{zhang2016degeneracy,hinduja2019degeneracy}, covariance
inflation and scalar down-weighting attenuate a sensor in all directions,
and scene-level quality gating~\cite{zheng2022fastlivo2} decides whether a
visual update is applied, not in which directions; their thresholds are
set by hand.

\emph{2) Computation on the platform.}
The processors of these platforms have no graphics processing unit (GPU)
acceleration and are shared with planning and control, and memory bounds
the mapped area.
The state update dominates the cost of an iterated LIVO filter.
Sequential designs add a second iterated update for vision; in
FAST-LIVO2~\cite{zheng2022fastlivo2} it runs over every patch pixel at
several pyramid levels, even when the LiDAR already constrains the pose.

To address these problems, this article presents SA-LIVO (Fig.~\ref{fig:system}), a real-time LIVO
system that admits the visual information in each direction only with the
confidence its error model supports.
The visual measurements are compressed once into an information form, and
the joint update iterates on an $18$-dimensional matrix whose cost does not
depend on their number.
The main contributions are as follows.

\begin{enumerate}
\item[1)] \textbf{Subspace-Aware Information Fusion (SAIF).}
  We model the error shared by all observations of a camera frame as a pose
  offset with an uncertainty floor and derive the exact fusion: the
  information of each visual direction saturates at a ceiling, weak
  directions keep theirs, and the LiDAR is never attenuated. The floor is
  estimated online, without a tuning threshold.

\item[2)] \textbf{An efficiency-oriented tightly coupled estimator.}
  Point-to-plane and photometric residuals are fused with the IMU prior in
  one invariant extended Kalman filter (InEKF) loop that touches no image
  pixel; window reuse and whole-image errors are modeled without tuning
  constants, and frozen, constant-time voxel planes keep the LiDAR
  residuals cheap and independent of the state.

\item[3)] \textbf{System implementation and validation.}
  SA-LIVO runs in real time on a laptop processor and an embedded ARM board
  without GPU, with a memory footprint well below that of FAST-LIVO2,
  R3LIVE, and SR-LIVO. On 24 sequences of HILTI'22~\cite{zhang2022hilti} and
  MARS-LVIG~\cite{li2024marslvig} it is the most accurate or tied on 22, and
  it is validated on self-collected handheld and unmanned aerial vehicle
  (UAV) data; code and data
  will be released.
\end{enumerate}

The rest of this article is organized as follows.
Section~\ref{sec:related} reviews related work,
Sections~\ref{sec:overview}--\ref{sec:update} present the method,
Section~\ref{sec:experiments} evaluates it, and
Section~\ref{sec:conclusion} concludes.

\section{Related Work}
\label{sec:related}

\subsection{LiDAR Odometry and Degeneracy}

Point-to-plane LiDAR odometry~\cite{zhang2014loam,xu2022fastlio2} is
accurate in structured scenes, and MSC-LIO~\cite{zhang2025msclio} brings the
efficiency of the multistate constraint Kalman filter to it through
same-plane cluster tracking.
Where the scan geometry leaves a direction unconstrained, as in corridors,
open floors, or tunnels, the matching Hessian becomes nearly singular.
Degeneracy-aware methods detect this from its eigenvalues and reject or
constrain the update~\cite{zhang2016degeneracy,hinduja2019degeneracy},
filter constraints direction-wise inside the registration~\cite{tuna2024xicp}, or
select an outlier-rejection mode and a GNSS-aided recovery~\cite{chen2026fastlivgo}.
They act on weak information through hand-set thresholds; SAIF instead
attenuates over-confident visual information by an error model estimated
online, in the spirit of innovation-based adaptive Kalman
filtering~\cite{mehra1972}.
For the map, we build on the voxel planes with uncertainty of
VoxelMap~\cite{yuan2022voxelmap}, whose planes keep absorbing returns
registered under the current, drifting estimate; we freeze saturated planes
and prefer them in the association.

\subsection{Visual-Inertial Odometry and Invariant Filtering}

\begin{figure*}[!t]
  \centering
  \includegraphics[width=0.95\textwidth]{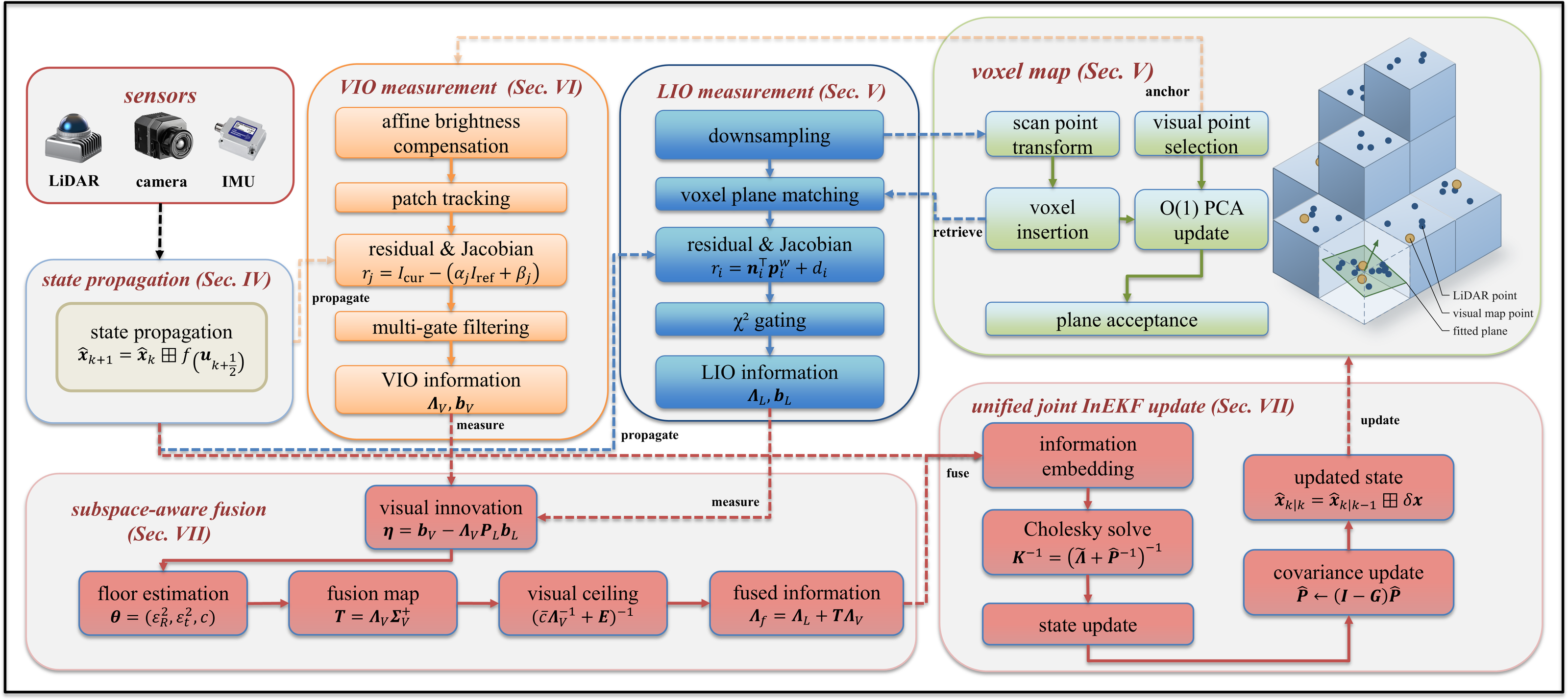}
  \caption{Overview of SA-LIVO. The LiDAR (\textbf{blue}) and visual
    (\textbf{orange}) paths each yield an information form; SAIF (\textbf{red})
    admits the visual information up to its ceiling and adds the LiDAR
    information for the joint invariant extended Kalman filter (InEKF)
    update; the corrected pose updates the
    voxel map (\textbf{green}). Dashed arrows: data flow between modules.}
  \label{fig:system}
\end{figure*}

Filter-based VIO descends from MSCKF~\cite{mourikis2007msckf};
ROVIO~\cite{bloesch2017rovio} applies
direct photometric updates in an iterated EKF.
Keyframe systems such as VINS-Mono~\cite{qin2018vins} triangulate their map points, whose depth
uncertainty grows with range.
Direct methods~\cite{engel2018dso} sample photometric
residuals at given 3D points and therefore suit LiDAR anchoring;
DSO~\cite{engel2018dso} also introduced per-frame affine brightness
correction with first-estimate Jacobians~\cite{huang2010fej,hesch2014vio}.
IR-VIO~\cite{song2023irvio} counters illumination change and blur with image
enhancement and adaptive weights chosen to maximize a confidence measure,
whereas our frame weight is the covariance-matching estimate of the frame's
noise.
The InEKF~\cite{barrau2017invariant} derives trajectory-independent
propagation Jacobians from group-affine symmetry and preserves the
unobservable subspace~\cite{brossard2019consistency}; it has been applied to
LiDAR-inertial odometry~\cite{shi2023invlio}, and
transformations between equivariant filters give consistent and efficient
visual-inertial navigation~\cite{tian2026eqf}.
We use its right-invariant attitude error for LiDAR and photometric
residuals in one update loop.

\subsection{LiDAR-Visual-Inertial Fusion}

Loosely coupled systems let one sensor initialize the
other; tightly coupled systems fuse the residuals of all sensors.
LIC-Fusion~\cite{zuo2019licfusion} fuses features in an MSCKF,
R$^2$LIVE~\cite{lin2021r2live} and FAST-LIVO~\cite{zheng2022fastlivo} use
direct photometric updates on LiDAR-anchored points, and
FAST-LIVO2~\cite{zheng2022fastlivo2} runs sequential iterated LiDAR and
visual updates, linearized at different states;
KILVO~\cite{gao2026kilvo} follows the same LiDAR-then-photometric sequence
on humanoid robots, with leg kinematics as an additional constraint.
R3LIVE~\cite{lin2022r3live} maintains a dense colored map at a higher
per-frame cost.
Smoothing-based systems include LVI-SAM~\cite{shan2021lvisam},
continuous-time fixed-lag smoothing~\cite{lv2023ctlic}, and
GIVL-SLAM~\cite{wang2025givl}, which adds GNSS carrier-phase observations.
REA-LVIO~\cite{wen2026realvio} updates sparse visual features and the
LiDAR asynchronously in an MSCKF, so that a degraded modality does not
corrupt the other.
In these designs the weighting between LiDAR and visual information is the
same in all pose directions and set by hand, not derived from a model of
the measurement errors (Table~\ref{tab:compare}).
SA-LIVO derives the fusion from explicit, online-estimated models of the
visual errors within one information form.

\begin{table}[!t]
  \centering
  \scriptsize
  \renewcommand{\arraystretch}{0.95}
  \setlength{\tabcolsep}{2.5pt}
  \caption{Comparison of Representative LiDAR-Visual-Inertial Systems}
  \label{tab:compare}
  \begin{tabular}{@{}lcccc@{}}
    \toprule
    Method & Estimator & Visual residual & LiDAR / visual & Visual weight \\
    \midrule
    LVI-SAM~\cite{shan2021lvisam} & Smoothing & Features & Two subsystems & Fixed \\
    GIVL-SLAM~\cite{wang2025givl} & Smoothing & Features & Joint & Fixed \\
    R3LIVE~\cite{lin2022r3live} & ESIKF & Photometric & Sequential & Fixed \\
    FAST-LIVO2~\cite{zheng2022fastlivo2} & ESIKF & Photometric & Sequential & Fixed \\
    KILVO~\cite{gao2026kilvo} & ESIKF & Photometric & Sequential & Mode switching \\
    REA-LVIO~\cite{wen2026realvio} & MSCKF & Features & Asynchronous & Fixed \\
    \textbf{SA-LIVO} & InEKF & Photometric & Joint & Online error model \\
    \bottomrule
  \end{tabular}
\end{table}

\section{System Overview}
\label{sec:overview}

Fig.~\ref{fig:system} shows the system; $i$ indexes LiDAR points and $j$
visual observations, and hats denote estimates.
LiDAR points are re-combined into scans at the camera rate (10~Hz) by IMU
(200~Hz) backward propagation.
The LiDAR branch (Section~\ref{sec:lio_constraint}) yields the information
form $\bLam_L \in \mathbb{R}^{6\times6}$, $\bb_L \in \mathbb{R}^6$ of the
gated point-to-plane residuals; the visual branch
(Section~\ref{sec:vio_constraint}) compresses the photometric residuals of
a sliding window of LiDAR-anchored templates, once at the propagated state,
into $\bLam_V \in \mathbb{R}^{18\times18}$, $\bb_V \in \mathbb{R}^{18}$.
SAIF (Section~\ref{sec:model}) fuses the two for one iterated InEKF update
(Section~\ref{sec:update}), and the voxel map is updated with the refined
pose.

\section{State Representation and IMU Propagation}
\label{sec:state_imu}
\label{sec:imu}

With the IMU frame as the body frame~\cite{xu2022fastlio2}, the state is
\begin{equation}
  \bx \trigeq \bigl[
    \bR^\top,\;
    \bp^\top,\;
    \bv^\top,\;
    \mathbf{b}_g^\top,\;
    \mathbf{b}_a^\top,\;
    \bg^\top
  \bigr]^\top \in \mathcal{M} \trigeq \SO \times \mathbb{R}^{15},
  \label{eq:state}
\end{equation}
where $\bR \in \SO$ is the IMU-to-world rotation, $\bp, \bv \in \mathbb{R}^3$
are the world-frame position and velocity, $\mathbf{b}_g, \mathbf{b}_a$ the
gyroscope and accelerometer biases, and $\bg$ the gravity vector, estimated
online.
With hats denoting estimates, the error state
$\delta\bx = \bx \bminus \bxhat \in \mathbb{R}^{18}$, whose inverse
$\bx = \bxhat \bplus \delta\bx$ applies a correction, uses the
right-invariant attitude error of the
InEKF~\cite{barrau2017invariant},
$\delta\boldsymbol{\phi} = \Log(\bR\hat{\bR}^\top)$, i.e.\
$\bR = \Exp(\delta\boldsymbol{\phi})\hat{\bR}$ with $\Exp$ and $\Log$ the
exponential and logarithm maps of $\SO$, while the errors of position,
velocity, biases, and gravity are Euclidean
($\delta\bp = \bp-\hat{\bp}$, etc.).
The attitude error then has an identity self-transition, which removes the
per-step rotation of the error that the left-invariant convention of
FAST-LIO2~\cite{xu2022fastlio2} incurs.
The Euclidean $\delta\bp$ and $\delta\bv$ keep the LiDAR and visual
information forms, and the fusion of Section~\ref{sec:model}, in one linear
error space; the price is the group-affine consistency of the full
SE$_2$(3) filter~\cite{barrau2017invariant,brossard2019consistency}, which
this hybrid form does not claim (supplementary Section~S-I-A).
Between sensor frames the state is propagated with the IMU measurements and
its covariance as
$\Phat \leftarrow \mathbf{F}_x\Phat\mathbf{F}_x^\top + \mathbf{F}_w\mathbf{Q}\mathbf{F}_w^\top$,
with $\mathbf{F}_x$, $\mathbf{F}_w$ the state and noise Jacobians of the
discrete kinematics of~\cite{xu2022fastlio2} and $\mathbf{Q}$ the IMU noise
covariance; the same propagation undistorts the LiDAR points.
$\Phat_{0:6}$ denotes the pose block of $\Phat$, and $\bP_R$, $\bP_t$ its
rotation and translation blocks.

\section{Drift-Resilient Multi-Scale LIO}
\label{sec:lio_constraint}

A residual enters $(\bLam_L,\bb_L)$ as a measurement whose error is
independent of the state error; the map and the plane association are built
so that this holds for the planes actually matched.

\subsection{Point-to-Plane Residual}
\label{sec:lio_residual}

A body-frame LiDAR point $\mathbf{p}^b$ is transformed with the LiDAR--IMU
extrinsic $(\bR_{LI}, \bp_{LI})$ and the state:
\begin{equation}
  \mathbf{p}^I = \bR_{LI}\mathbf{p}^b + \bp_{LI}, \qquad
  \mathbf{p}^w  = \bR\mathbf{p}^I + \bp.
  \label{eq:coordchain}
\end{equation}
Matched to a plane with unit normal $\bn_i$ and centroid $\mathbf{m}_i$, the
point has the residual and, with respect to
$[\delta\boldsymbol{\phi};\delta\bp]$, the Jacobian
\begin{equation}
  r_i = \bn_i^\top(\mathbf{p}^w_i - \mathbf{m}_i),\qquad
  \mathbf{h}_i = \begin{bmatrix}
    \skewop{\bR\mathbf{p}^I_i}\bn_i \\
    \bn_i
  \end{bmatrix} \in \mathbb{R}^{6},
  \label{eq:residual}
\end{equation}
where $\skewop{\cdot}$ is the skew-symmetric matrix; the rotation block
follows from $\bR \leftarrow \Exp(\delta\boldsymbol{\phi})\bR$ and depends on
the world-centered vector $\bR\mathbf{p}^I_i$ (supplementary Section~S-II-A).

\subsection{Voxel Map and Plane Freezing}
\label{sec:lio_map}

The map is a hash table of root voxels of side $v_s$, each holding an
octree~\cite{yuan2022voxelmap} with at most two subdivision levels.
A plane is fitted to the returns of a voxel by principal component analysis
(PCA), with eigenvalues $\lambda_{\min}\le\lambda_{\rm mid}\le\lambda_{\max}$
of their scatter matrix; a voxel whose fit is not planar is split, down to
side $v_s/4$.
A fit is accepted as planar when
\begin{equation}
  \lambda_{\min} < \varepsilon_{\rm pl}.
  \label{eq:planar_test}
\end{equation}
$\lambda_{\min}$ is the mean squared distance of the support to the plane,
so $\sqrt{\varepsilon_{\rm pl}}$ is the admissible RMS thickness of the
surface in meters, the quantity the residual~\eqref{eq:residual} ignores;
$\varepsilon_{\rm pl}$ is set per sensor.

\begin{figure}[!t]
  \centering
  \includegraphics[width=\columnwidth]{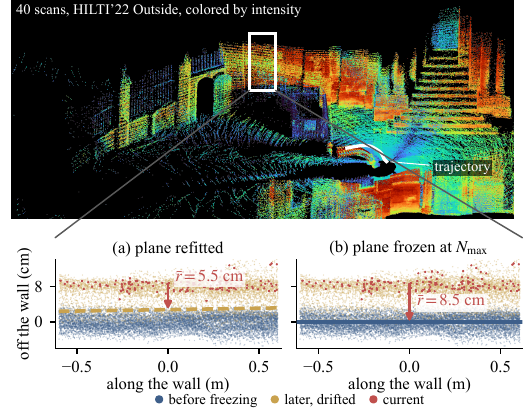}
  \caption{Plane freezing on real LiDAR data (HILTI'22 Outside, 40 scans,
    raw returns colored by intensity). After the voxels of the marked wall segment reach
    the freezing count $N_{\max}$, a sudden drift of 8~cm along the wall normal is added
    to the later scans. (a)~A plane refitted on all returns is pulled toward
    the drifted returns and hides part of the drift from the current scan.
    (b)~The plane frozen at $N_{\max}$ returns keeps the full residual.}
  \label{fig:lio_map}
\end{figure}

An accepted plane is refitted as returns arrive and is \emph{frozen} once
its voxel holds $N_{\max}$ returns (Fig.~\ref{fig:lio_map}): its parameters are fixed and its points
released.
Every return is registered under the pose estimate of its scan, so a plane
inherits the registration errors of the scans that built it.
A plane built over many scans averages errors that are decorrelated by the
time it is matched again, so its error is nearly independent of the
current state error, as the information form~\eqref{eq:LamL_final} assumes.
Returns inserted later are registered under the current, drifting estimate;
refitting on them would move the plane with the drift and correlate its
error with the error it is meant to correct.
Beyond $N_{\max}$ returns these correlated errors leave little precision to
gain (Section~\ref{sec:lio_noise}), so freezing loses little information;
it also bounds the memory and the refitting cost of a voxel.

\subsection{\texorpdfstring{$O(1)$}{O(1)} Neighborhood PCA}
\label{sec:lio_pca}

Where the plane of a voxel is missing or still unreliable, a coarser fit
over its neighbors is needed.
Each root voxel keeps the sufficient statistics of all returns
$\{\mathbf{x}_k\}_{k=1}^{n}$ inserted into it,
\begin{equation}
  \mathbf{s}_1 = \textstyle\sum_{k}\mathbf{x}_k,\quad
  \mathbf{S}_2 = \textstyle\sum_{k}\mathbf{x}_k\mathbf{x}_k^\top,\quad
  n,
  \label{eq:ss}
\end{equation}
from which the scatter matrix is
$\mathbf{S}_2/n - (\mathbf{s}_1/n)(\mathbf{s}_1/n)^\top$.
A neighborhood of half-width $\ell$ voxels is fitted with $(2\ell+1)^3$
additions and one $3\times3$ eigendecomposition, whatever the number of
points.
The statistics keep accumulating after a plane is frozen, so the
neighborhood fit describes the geometry as currently registered.

\subsection{Noise Model and LiDAR Information}
\label{sec:lio_noise}
\label{sec:lio_infomat}

Each point carries the body-frame covariance $\bSig_b$ of the
range--bearing model of~\cite{yuan2022voxelmap}, propagated to the world
frame together with the state uncertainty:
\begin{equation}
\begin{gathered}
  \bSig_w = \bSig_w^{s}
           + \skewop{\bR\mathbf{p}^I}\bP_R\skewop{\bR\mathbf{p}^I}^\top + \bP_t,\\
  \bSig_w^{s} = \bR\bR_{LI}\bSig_b\bR_{LI}^\top\bR^\top,
\end{gathered}
  \label{eq:worldcov}
\end{equation}
with the world-centered skew that the right-invariant Jacobian requires.
Each plane carries a $6\times6$ covariance $\bSig_{\rm plane}$ of
$(\bn,\mathbf{m})$.
For a voxel plane it is propagated from the per-point $\bSig_w$ through the
PCA by first-order eigenvector perturbation~\cite{yuan2022voxelmap}.
A neighborhood fit keeps no per-point covariances and uses the standard
error of a least-squares plane,
\begin{equation}
  \bSig_{\rm plane} = \frac{\lambda_{\min}}{N_{\rm eff}}\,\bI_6,
  \qquad
  N_{\rm eff} = \min(N,\,N_{\max}),
  \label{eq:planecov}
\end{equation}
with $N$ the number of aggregated returns: a thick fit is uncertain however
many points support it, a fit from few points however thin it is.
The count is capped because the returns of a region share a registration
error: for $N$ samples with pairwise correlation $\varrho$ the effective
sample size $N/(1+(N-1)\varrho)$ saturates at $1/\varrho$, and $N_{\rm eff}$
is its piecewise-linear bound with $N_{\max}\approx1/\varrho$, the count at
which a voxel plane freezes (supplementary Section~S-II-B).

The residual variance is
\begin{equation}
\begin{gathered}
  \sigma_i^2 = \mathbf{J}_{nm,i}\bSig_{\rm plane}\mathbf{J}_{nm,i}^\top
             + \bn_i^\top\bSig_{w,i}^{s}\bn_i + \varepsilon_0,\\
  \mathbf{J}_{nm,i} = \bigl[(\mathbf{p}^w_i-\mathbf{m}_i)^\top,\;-\bn_i^\top\bigr],
\end{gathered}
  \label{eq:sigmatot}
\end{equation}
with $\varepsilon_0 = 10^{-3}$\,m$^2$ for unmodeled roughness.
Only the sensor term $\bSig_w^{s}$ enters, since the state uncertainty is
carried by the prior in the joint solve (Section~\ref{sec:update}).
A residual with $r_i^2/\sigma_i^2 > g_L^2$ is rejected ($g_L$ standard
deviations), and the set $\mathcal{I}_L$ of the others gives
\begin{equation}
  \bLam_L = \textstyle\sum_{i\in\mathcal{I}_L}\sigma_i^{-2}\mathbf{h}_i\mathbf{h}_i^\top,
  \qquad
  \bb_L   = -\textstyle\sum_{i\in\mathcal{I}_L}\sigma_i^{-2}\mathbf{h}_i r_i .
  \label{eq:LamL_final}
\end{equation}
The matched planes do not change across InEKF iterations, so the
correspondences are cached after the first iteration and later iterations
recompute only the distances $r_i$.

\subsection{Saturation-Priority Multi-Scale Plane Association}
\label{sec:lio_satpri}

Each point is matched at two scales.
\emph{Scale-0} is the plane in the octree of the point's voxel with the
largest likelihood $\sigma^{-1}\exp(-r_i^2/2\sigma^2)$, $\sigma^2$ being
the plane and point variance along the normal.
\emph{Scale-1} is the neighborhood fit of Section~\ref{sec:lio_pca} with
$\ell = \mathrm{clamp}(\lceil r_s/v_s\rceil, 1, 5)$ for a support radius
$r_s$, accepted by~\eqref{eq:planar_test} when it holds at least
$\max(10,(2\ell+1)^2)$ returns.
A plane is used only if the point lies within three plane radii
$\sqrt{\lambda_{\max}}$ of its centroid laterally.
When both scales return a plane, candidates whose normals deviate by more
than $18^\circ$ from that of the thinnest one are discarded, thinness being
compared by the scale-free ratio
\begin{equation}
  \mu = \frac{\lambda_{\min}}{\lambda_{\min}+\lambda_{\rm mid}+\lambda_{\max}}
  \in \left[0,\tfrac{1}{3}\right],
  \label{eq:kappa}
\end{equation}
since $\lambda_{\min}$ grows with the support and $\mu$ does not.
The remaining candidates are ranked lexicographically: frozen planes
first, then thin planes ($\mu < 0.15$), then the smallest $|r_i|/\sigma_i$.
A plane that is still being refitted, and the neighborhood fit, include
returns registered under the current estimate, so their residual depends
on the state error through both the point and the plane and pulls the
estimate toward its own drift; a frozen plane does not.
The same correlation is carried exactly for the visual templates
(Section~\ref{sec:sw}); the LiDAR map holds orders of magnitude more planes,
so the association instead selects the matches for which the correlation is
negligible, and uses the most recent fit only where no frozen plane exists.
The association decides which plane a point is matched to, never how much
the match weighs, and its constants are geometric and the same for all
sequences (details in supplementary Section~S-II-C).

\section{LiDAR-Anchored Direct Photometric VIO}
\label{sec:vio_constraint}

Visual map points are LiDAR points~\cite{zheng2022fastlivo2}: each is a
world-frame point $\mathbf{p}^w_j$ with the covariance $\bSig_{w,j}$
of~\eqref{eq:worldcov} and the normal of its voxel plane, so no depth is
triangulated.
With the IMU-to-camera extrinsic $(\bR_{ci},\bp_{ci})$, a point maps to the
camera frame as
$\mathbf{p}^c_j = \bR_{ci}\bR^\top(\mathbf{p}^w_j - \bp) + \bp_{ci}$
and to the image by the camera model $\pi$.
Map-point management, the selection of at most one point per image cell by
gradient energy, and the depth-continuity and view-angle filters follow
FAST-LIVO2~\cite{zheng2022fastlivo2}, with its thresholds.

\subsection{Affine Brightness Model}
\label{sec:affine}

The brightness transfer between a reference template $I_{\rm ref}$ and the
current image $I_{\rm cur}$ is modeled as
\begin{equation}
  I_{\rm cur} \approx \alpha_j\,I_{\rm ref} + \beta_j,
  \label{eq:affine}
\end{equation}
where $(\alpha_j,\beta_j)$ are the parameters of the window frame that
holds the reference template of observation $j$, shared by all its
observations.
They are fitted by gradient-weighted least squares on the patch means and
clamped to $\alpha_j\in[0.3,3]$, $\beta_j\in[-100,100]$ gray levels
(supplementary Section~S-III-A).
The affine transfer is a first-order description of the camera response,
and its error grows with the exposure ratio; beyond a ratio
$e^{\delta_\alpha}$ ($\delta_\alpha=0.5$) the current-image noise is
therefore inflated by $1/w_{\rm cam}$, with
$w_{\rm cam} = \exp\bigl(-\max(|\log\alpha_j|-\delta_\alpha,0)\bigr)$.

\subsection{Photometric Residual and Pre-Loop Jacobian}
\label{sec:fej}

Observation $j$ samples a $3\times3$ pattern of taps at $3$~px spacing
around its projection $\pi(\mathbf{p}^c_j)$, bilinearly interpolated; its
residual is the mean of the nine affine-compensated differences
\begin{equation}
  r_j = \frac{1}{9}\sum_{k=1}^{9}
    \bigl[I_{\rm cur}(\mathbf{u}_k)
    - \alpha_j\,I_{\rm ref}(\mathbf{u}_k^{\rm ref}) - \beta_j\bigr],
  \label{eq:photo_res}
\end{equation}
with $\mathbf{u}_k$ and $\mathbf{u}_k^{\rm ref}$ the tap locations in the
current image and in the reference template, the latter fixed when the
observation is created.
Under $\bR \leftarrow \Exp(\delta\boldsymbol{\phi})\bR$ its Jacobian with
respect to $[\delta\boldsymbol{\phi};\delta\bp]$ is (supplementary
Section~S-III-B)
\begin{equation}
  \mathbf{J}_j = \nabla I_j
    \begin{bmatrix}
      \skewop{\mathbf{p}^c_j - \bp_{ci}}\bR_{ci}\bR^\top & -\bR_{ci}\bR^\top
    \end{bmatrix}
  \in \mathbb{R}^{1\times6},
  \label{eq:photo_jac}
\end{equation}
where $\nabla I_j = [I_x,I_y]\,\mathbf{J}_\pi \in \mathbb{R}^{1\times3}$
composes the image gradient $[I_x,I_y]$ at the projection with the
Jacobian $\mathbf{J}_\pi \in \mathbb{R}^{2\times3}$ of $\pi$.
The noise variance of the observation is
\begin{equation}
  \sigma_{j}^2 = \sigma_{\rm px}^2 + \nabla I_j\bSig_{c,j}\nabla I_j^\top,
  \qquad
  \bSig_{c,j} = \bR_{ci}\bR^\top\bSig_{w,j}\bR\bR_{ci}^\top,
  \label{eq:photo_noise}
\end{equation}
with $\sigma_{\rm px}$ the photometric noise of the camera in gray levels, and the second term the uncertainty of the anchor point.

The visual information form is assembled once at the propagated state
$\bxhat^0$ and held across the InEKF iterations, while the LiDAR residuals
are re-evaluated at every iterate (Section~\ref{sec:update}).
A photometric residual is linear in the pose only within about one patch
radius of its linearization point, a point-to-plane residual over a much
wider range.
With $\bb_V$ held, the stationary points of the iteration minimize the LiDAR
cost exactly and the photometric cost through its first-order model at
$\bxhat^0$, and $\bLam_V$ acts as a Levenberg--Marquardt damping metric
along strongly constrained visual directions (supplementary
Section~S-I-C); re-centering $\bb_V$ would extrapolate the photometric
model to the displacements produced by the LiDAR.
The posterior covariance uses the full $\bLam_V$.
At the convergence threshold $\epsilon_\phi = 0.01^\circ$ a projection moves
by about $0.09$~px per iterate (focal length $500$~px), so one evaluation of
$\mathbf{J}_j$ saves up to four gradient and projection evaluations per
observation without measurable change.
Unlike first-estimate Jacobians~\cite{huang2010fej,hesch2014vio}, this is
confined to one update, and no iteration touches a pixel
(Fig.~\ref{fig:update_loop}).

\subsection{Sliding Window and Correlated Template Errors}
\label{sec:sw}

\begin{figure}[!t]
  \centering
  \includegraphics[width=\columnwidth]{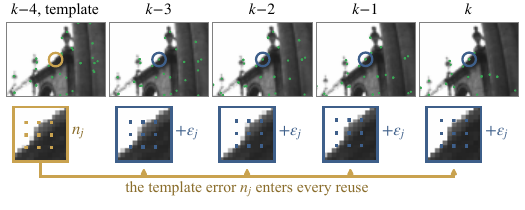}
  \caption{Reuse of one visual observation in the sliding window
    (HILTI'22 Outside; green: the window observations of our system). Its
    $3\times3$ template is sampled once, at update $k\!-\!4$, and compared
    with the current image in the next four updates, each adding new image
    noise $\epsilon_j$; the shared template error $n_j$ correlates these
    updates and is carried by the cross-covariance $\mathbf{c}_j$.}
  \label{fig:sw_reuse}
\end{figure}

The window $\mathcal{W}$ holds the $W = 5$ most recent frames ($0.5$~s at
$10$~Hz), each with at most $N_{\rm obs}$ observations; a frame that moves
less than $\delta_R$ in rotation and $\delta_t$ in translation from the
previous one replaces it.
Every observation is evaluated in the current image, and successive frames
admit largely different points, so the window holds up to
$WN_{\rm obs}$ points spread over the scene and raises the rank and the
conditioning of $\bLam_V$.

The price is reuse (Fig.~\ref{fig:sw_reuse}): the template error of an
observation enters every update that reuses it.
Writing $y_j \trigeq -r_j$ and linearizing at the prior,
\begin{equation}
  y_j = \mathbf{J}_j\,\delta\bx_{0:6} + n_j + \epsilon_j,
  \label{eq:template_model}
\end{equation}
where $\delta\bx_{0:6}$ is the pose part of the error state, $n_j$ the error
of the reference template, common to every reuse, and $\epsilon_j$ the noise
of the current image, new at every update.
Both images come from the same camera, so the pixel noise $\sigma_{\rm px}^2$
splits equally between them: $\epsilon_j$ has variance
$\sigma^2_{\epsilon,j}/w_j$ with $\sigma^2_{\epsilon,j}=\tfrac12\sigma_{\rm px}^2$
and the robust weight $w_j$ of Section~\ref{sec:vio_info}, and $n_j$ has
variance $\sigma^2_{n,j} = \sigma_j^2 - \tfrac12\sigma_{\rm px}^2$, which
includes the anchor term of~\eqref{eq:photo_noise}.

Following the Schmidt (consider) filter~\cite{schmidt1966}, the template
errors are not estimated, but their cross-covariances with the state,
$\mathbf{c}_j \trigeq \mathrm{Cov}(\delta\bx, n_j) \in \mathbb{R}^{18}$, are
propagated exactly: $\mathbf{c}_j = \mathbf{0}$ when the observation is
created, $\mathbf{c}_j \leftarrow \boldsymbol{\Phi}\mathbf{c}_j$ with the
IMU transition $\boldsymbol{\Phi}$ between updates (the product of the
$\mathbf{F}_x$ of Section~\ref{sec:state_imu}), and
$\mathbf{c}_j \leftarrow \mathbf{A}\mathbf{c}_j - \sigma^2_{n,j}\mathbf{k}_j$
after each update, with $\mathbf{A}$ the coefficient of the prior error in
the posterior error and $\mathbf{k}_j$ the gain of observation $j$
(Section~\ref{sec:bias_floor}).
Conditioning $n_j$ on the state error gives the measurement model of the
update,
\begin{equation}
\begin{gathered}
  \mathbf{y} = \tilde{\bH}\,\delta\bx + \boldsymbol{\nu},\qquad
  \tilde{\bH} = [\mathbf{J}\;\;\mathbf{0}] + \mathbf{C}^\top\Phat^{-1},\\
  \mathrm{Cov}(\boldsymbol{\nu}) = \tilde{\bSig} = \mathbf{D} - \mathbf{C}^\top\Phat^{-1}\mathbf{C},
\end{gathered}
  \label{eq:schmidt}
\end{equation}
where $\mathbf{y}$ and $\mathbf{J}$ stack the $y_j$ and $\mathbf{J}_j$ of the
$N$ accepted observations, $[\mathbf{J}\;\;\mathbf{0}]$ pads $\mathbf{J}$
with zeros for the twelve non-pose states,
$\mathbf{C} = [\mathbf{c}_1,\ldots,\mathbf{c}_N]$, and
$\mathbf{D} = \diag(d_j)$ with $d_j = \sigma^2_{n,j} + \sigma^2_{\epsilon,j}/w_j$.
Since $\mathbf{C}^\top\Phat^{-1}\mathbf{C}$ has rank at most $18$,
$\tilde{\bSig}^{-1}$ follows from the Woodbury identity in $O(N\cdot18^2)$.
The update equals that of an augmented-state Schmidt filter that carries
every template error (supplementary Section~S-III-C), at the cost of an
$18$-vector per observation, and since $\mathbf{c}_j$ spans all $18$
states, the visual information acts on velocity, biases, and gravity as
well.
No discount of reused observations and no covariance inflation is needed.

\subsection{Frame-Level Noise Scale}
\label{sec:vio_quality}

Motion blur, exposure changes the affine model misses, reflections, and
moving objects inflate the residuals of many observations of a frame at
once.
We treat them as a frame-level scale $\kappa$ of the modeled variance and
estimate it by covariance matching~\cite{mehra1972} over the
frame's candidate observations, i.e.\ before the outlier gates of
Section~\ref{sec:vio_info}: from
$\mathbb{E}[r_j^2] = \mathbf{J}_j\Phat_{0:6}\mathbf{J}_j^\top + \kappa\,\sigma_j^2$,
\begin{equation}
  \hat{\kappa} = \frac{\sum_j \bigl(r_j^2 - \mathbf{J}_j\Phat_{0:6}\mathbf{J}_j^\top\bigr)}
                      {\sum_j \sigma_j^2},
  \qquad
  q = \frac{1}{\max(1,\hat{\kappa})}.
  \label{eq:q_vio}
\end{equation}
Unmodeled errors add to the modeled noise and never reduce it, so the
frame's measurement covariance is scaled by $\max(1,\hat\kappa)$, i.e.\ its
visual information by $q$.
The candidates are used on purpose: the gates truncate the residuals, and
the accepted residuals of a degraded frame look individually consistent
while their errors are correlated.
$q$ has no tuning constant, and as $q\to0$ the update degrades to a
LiDAR-only one without mode switching; with $q$ the visual information
fails a $\chi^2$ consistency test on 1--2\,\% of the frames, against
13--29\,\% without it (supplementary Section~S-V-B).

\subsection{Visual Information Matrix}
\label{sec:vio_info}

Observations enter only after the validity and outlier gates of
FAST-LIVO2~\cite{zheng2022fastlivo2} (depth, image bounds, view angle,
parallax, and a patch RMS residual below $\delta_{\rm abs}$) and a $\chi^2$
gate on $z_j = r_j/\sigma_j$ decided per frame.
Since $q$ shows that the noise of a frame departs from its model, the fixed
threshold on $z_j$ is replaced by the minimum-error decision of a mixture
fitted to the frame's $|z_j|$ by expectation
maximization: inliers $z_j\sim\mathcal{N}(0,s^2)$ with
unknown scale $s$, and outliers uniform on $[0,Z]$ with weight $\xi$, $Z$
being the largest $|z_j|$ of the frame.
An observation is accepted if
\begin{equation}
  |z_j| \leq g = s\,\sqrt{2\ln\frac{2(1-\xi)Z}{\xi\, s\sqrt{2\pi}}},
  \label{eq:adaptive_gate}
\end{equation}
with $g$ held in $[2s,\,5s]$ and a fixed gate $\tau_V$ for frames with fewer
than $20$ candidates (supplementary Section~S-III-D); the gate selects
but does not weight.
Accepted observations carry the robust weight $w_j = w_{\rm hub}\,w_{\rm cam}$,
with the Huber factor
$w_{\rm hub} = \min(1,\tau_{\rm hub}/|r_j|)$, $\tau_{\rm hub}=\delta_{\rm abs}/2$,
and $w_{\rm cam}$ of Section~\ref{sec:affine}.
Over the set $\mathcal{I}_V$ of the $N$ accepted observations, the visual
information form is
\begin{equation}
  \bLam_V = q\,\tilde{\bH}^\top\tilde{\bSig}^{-1}\tilde{\bH}
           \in \mathbb{R}^{18\times18},
  \qquad
  \bb_V   = q\,\tilde{\bH}^\top\tilde{\bSig}^{-1}\mathbf{y},
  \label{eq:LamV_final}
\end{equation}
which for $\mathbf{C} = \mathbf{0}$ reduces to
$q\sum_{j\in\mathcal{I}_V}\mathbf{J}_j^\top\mathbf{J}_j/d_j$ on the pose block.

\section{Subspace-Aware Information Fusion}
\label{sec:model}
\label{sec:icsf}

Adding the LiDAR and visual information forms treats both models as exact.
After the frame scale $q$ and the template cross-covariances, one visual
error remains: an extrinsic or time-offset error, rolling shutter, or
template drift displaces all projected points of a frame consistently, a
pose error of the whole frame that does not average out
(Fig.~\ref{fig:saif_concept}(c)), so vision is most over-confident where it
is strongest.
SAIF models this frame-common error, derives the exact fusion under it, and
estimates it online.

The LiDAR information lives on the pose block of $\delta\bx$
($\mathbf{h}_i$ of~\eqref{eq:residual} has no other component);
$\bar{\bLam}_L$, $\bar{\bb}_L$ denote its embedding in the $18$-dimensional
state.
SAIF attenuates only the visual information: the plane model carries a
calibrated covariance (Section~\ref{sec:lio_noise}), and an inconsistency of
the LiDAR residuals appears in the strongly constrained directions, pointing
to the motion prior rather than to the measurement.

\subsection{Frame-Common Error Model}
\label{sec:formulation}

At the prior mean, the visual information vector is modeled as
\begin{equation}
\begin{gathered}
  \bb_V = \bLam_V\bigl(\delta\bx + \boldsymbol{\delta}\bigr) + \mathbf{w},\qquad
  \mathbf{w}\sim\mathcal{N}(\mathbf{0},\,c\,\bLam_V),\\
  \boldsymbol{\delta}\sim\mathcal{N}(\mathbf{0},\mathbf{E}),\qquad
  \mathbf{E} = \diag\bigl(\varepsilon_R^2\bI_3,\;\varepsilon_t^2\bI_3,\;\mathbf{0}_{12}\bigr).
\end{gathered}
  \label{eq:bias_model}
\end{equation}
$\mathbf{w}$ is the photometric noise of the individual observations, with
$c$ the noise scale left after $q$; $\boldsymbol{\delta}$ is the
frame-common error, a pose offset in rotation (rad) and translation (m)
shared by all observations of the frame and independent of the LiDAR and of
the state error.
$\varepsilon_R$ and $\varepsilon_t$ are the \emph{uncertainty floor} of the
visual information: no number of observations constrains the pose of a
frame better than $\mathbf{E}$.
$\mathbf{E}$ is isotropic within the rotation and within the translation
block, so the model is stated in consistent units and does not depend on
the orientation of the world frame.
For $\mathbf{E}=\mathbf{0}$ and $c=1$ it is the model of Section~\ref{sec:sw}.
Sources of such a floor are extrinsic calibration errors, a
synchronization error of about one millisecond, and the depth error of the
anchor points; the model does not need to separate them.

\subsection{Effective Information and Exact Fusion}
\label{sec:spectral_opt}
\label{sec:fusion}

\emph{Proposition.}
Under~\eqref{eq:bias_model}, with the LiDAR information exact and
independent of the visual one, the posterior of $\delta\bx$ given the prior
$\mathcal{N}(\mathbf{0},\Phat)$ and the two information forms is that of the
information-form update with
\begin{equation}
\begin{gathered}
  \bLam_f = \bar{\bLam}_L + \mathbf{T}\bLam_V,\qquad
  \bb_f = \bar{\bb}_L + \mathbf{T}\bb_V,\\
  \mathbf{T} = \bLam_V\bSig_V^{+},\qquad
  \bSig_V = \bar{c}\,\bLam_V + \bLam_V\mathbf{E}\bLam_V ,
\end{gathered}
  \label{eq:fusion}
\end{equation}
with $\bar c = c$ and $(\cdot)^{+}$ the pseudo-inverse; the fusion uses
$\bar{c}=\max(c,1)$, as explained below (proof in supplementary
Section~S-IV-A).

Writing $\bLam_V = \mathbf{L}\mathbf{L}^\top$ with $\mathbf{L}$ of full
column rank, the visual information that enters the update is
\begin{equation}
  \mathbf{T}\bLam_V = \mathbf{L}\bigl(\bar{c}\,\bI + \mathbf{L}^\top\mathbf{E}\mathbf{L}\bigr)^{-1}\mathbf{L}^\top
  \;=\; \bigl(\bar{c}\,\bLam_V^{-1} + \mathbf{E}\bigr)^{-1},
  \label{eq:eff_info}
\end{equation}
the second form when $\bLam_V$ is invertible: the photometric uncertainty
$\bar{c}\bLam_V^{-1}$, which shrinks as observations accumulate, and the
frame-common uncertainty $\mathbf{E}$, which does not, add in series.
Four properties follow.
(i)~For $\mathbf{E} = \mathbf{0}$ and $\bar{c}=1$, $\mathbf{T}$ projects
onto the range of $\bLam_V$ and the fusion is the standard information sum.
(ii)~For an eigenvector $\mathbf{u}_k$ of $\bLam_V$ with eigenvalue
$\lambda_k$ inside the rotation (or translation) block,
\begin{equation}
  \mathbf{u}_k^\top\mathbf{T}\bLam_V\mathbf{u}_k = \gamma_k\lambda_k,\qquad
  \gamma_k = \frac{1}{\bar{c} + \lambda_k\varepsilon^2},
  \label{eq:srif_gate}
\end{equation}
with $\varepsilon^2$ the floor of that block, so weak visual directions
($\lambda_k\varepsilon^2 \ll \bar{c}$) keep their information and strong
ones saturate smoothly at $1/\varepsilon^2$: the attenuation falls on the
directions in which vision is \emph{most} confident, the opposite of a
degeneracy threshold.
(iii)~The LiDAR is never attenuated: where it is degenerate, vision
contributes at most its ceiling and the IMU prior carries the rest.
(iv)~$\mathbf{T}\bLam_V$ is symmetric positive semidefinite and continuous
in $(c,\mathbf{E})$, so $\bLam_f$ is a valid information matrix without
regularization.
The fusion weighs each direction of the subspace spanned by the visual
information against its ceiling, which is the sense in which it is
subspace-aware.

The fusion uses $\bar{c}=\max(c,1)$, so the visual information is never
raised above its nominal value.
The residuals that reach $\bLam_V$ have passed the $\chi^2$ gate and the
Huber weight, which lower their second moment, to $0.77$ of the untruncated
value at the lowest gate $g=2s$ (supplementary Section~S-IV-B); a fitted
$c<1$ reflects this selection rather than a smaller noise.

\subsection{Online Estimation of the Frame-Common Error}
\label{sec:bias_floor}

\begin{figure}[!htbp]
  \centering
  \includegraphics[width=\columnwidth]{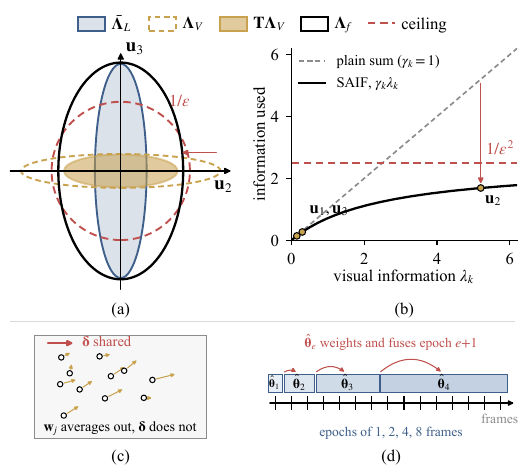}
  \caption{SAIF schematic (illustrative values). (a)~Information ellipses
    (semi-axes: square root of the information) in the plane of
    $\mathbf{u}_2$, constrained mainly by vision, and $\mathbf{u}_3$, by the
    LiDAR: the visual ellipse is pulled inside the ceiling $1/\varepsilon$,
    the LiDAR one is untouched. (b)~Visual information used along
    $\mathbf{u}_k$: weak directions keep it, the strong $\mathbf{u}_2$
    saturates at $1/\varepsilon^2$. (c)~The ceiling: the noise
    $\mathbf{w}_j$ of each projection averages out, the frame-common offset
    $\boldsymbol{\delta}$ does not. (d)~Online estimate: the
    $\hat{\boldsymbol{\theta}}$ of an epoch weights and fuses the next.}
  \label{fig:saif_concept}
\end{figure}

The parameters $\boldsymbol{\theta} = (\varepsilon_R^2,\varepsilon_t^2,c)$
are estimated from the visual innovation, the disagreement of the visual
information with the prior and the LiDAR.
With $\mathbf{P}_L = (\Phat^{-1} + \bar{\bLam}_L)^{-1}$, the innovation
\begin{equation}
  \boldsymbol{\eta} = \bb_V - \bLam_V\mathbf{P}_L\bar{\bb}_L
  \label{eq:vis_innov}
\end{equation}
has under~\eqref{eq:bias_model} a covariance that is linear in
$\boldsymbol{\theta}$,
\begin{equation}
\begin{gathered}
  \mathbf{S}(\boldsymbol{\theta}) = \mathbf{S}_0
    + \varepsilon_R^2\mathbf{S}_R + \varepsilon_t^2\mathbf{S}_t + c\,\mathbf{S}_c,\qquad
  \mathbf{S}_0 = \bLam_V\mathbf{P}_L\bLam_V,\\
  \mathbf{S}_R = \bLam_V\boldsymbol{\Pi}_R\bLam_V,\quad
  \mathbf{S}_t = \bLam_V\boldsymbol{\Pi}_t\bLam_V,\quad
  \mathbf{S}_c = \bLam_V,
\end{gathered}
  \label{eq:innov_cov}
\end{equation}
with $\boldsymbol{\Pi}_R$, $\boldsymbol{\Pi}_t$ the selectors of the
rotation and translation blocks
($\mathbf{E} = \varepsilon_R^2\boldsymbol{\Pi}_R + \varepsilon_t^2\boldsymbol{\Pi}_t$).
Over the frames $\tau$, $\boldsymbol{\theta}$ solves the weighted moment
equations~\cite{mehra1972}
\begin{equation}
  \textstyle\sum_\tau \tr\bigl(\mathbf{W}_\tau\mathbf{S}_{i,\tau}\mathbf{W}_\tau
  \bigl(\boldsymbol{\eta}_\tau\boldsymbol{\eta}_\tau^\top - \mathbf{S}_\tau(\boldsymbol{\theta})\bigr)\bigr) = 0,
  \quad i\in\{R,t,c\},
  \label{eq:moments}
\end{equation}
which are linear in $\boldsymbol{\theta}$, unbiased for any weight
$\mathbf{W}_\tau$ that does not depend on $\boldsymbol{\eta}_\tau$, and the
score equations of the Gaussian likelihood for
$\mathbf{W}_\tau = \mathbf{S}_\tau(\boldsymbol{\theta})^{+}$;
they are solved under $\boldsymbol{\theta}\geq\mathbf{0}$.
Past frames cannot be re-weighted without storing them, so the frames are
grouped into epochs of $1,2,4,\ldots$ frames, each weighted by the estimate
of the previous epoch (feasible generalized least squares,
Fig.~\ref{fig:saif_concept}(d)); a frame is fused with the estimate of the
last completed epoch, so its own innovation does not decide its weight.
The estimator keeps running sums only and has no window length, forgetting
factor, or threshold.
The terms of~\eqref{eq:innov_cov} vary differently across frames, which
separates the floor from the noise scale and from a mis-scaled prior
(supplementary Section~S-IV-C).

The template cross-covariances of Section~\ref{sec:sw} follow the
attenuated update.
With the posterior covariance $\mathbf{P}^+ = (\Phat^{-1}+\bLam_f)^{-1}$,
the coefficient of the prior error in the posterior error is
$\mathbf{A} = \bI - \mathbf{P}^+\bar{\bLam}_L
 - q\,\mathbf{P}^+\mathbf{T}\,\tilde{\bH}^\top\tilde{\bSig}^{-1}[\mathbf{J}\;\;\mathbf{0}]$,
and the gain of observation $j$ is
$\mathbf{k}_j = q\,\mathbf{P}^+\mathbf{T}\,\bigl[\tilde{\bH}^\top\tilde{\bSig}^{-1}\bigr]_{:,j}$,
the column of $\tilde{\bH}^\top\tilde{\bSig}^{-1}$ that multiplies $y_j$;
for $\mathbf{T} = \bI$ they are the Schmidt-filter gains.

\begin{algorithm}[!t]
\caption{Joint InEKF Update with SAIF}
\label{alg:inekf}
\small
\begin{algorithmic}[1]
\REQUIRE prior $(\bxhat^0, \Phat)$; scan; image; window $\mathcal{W}$;
         $\hat{\boldsymbol{\theta}}$ of the last completed epoch
\STATE $(\bLam_V, \bb_V)$ by~\eqref{eq:LamV_final};\;
       $\mathbf{T}$ from $\hat{\boldsymbol{\theta}}$ by~\eqref{eq:fusion}
\FOR{$t = 0, 1, \ldots$ until converged twice or $t = K_{\max}$}
  \STATE $(\bLam_L, \bb_L)$ at $\bxhat^t$ by~\eqref{eq:LamL_final};\;
         $\bLam_f$, $\bb_f$ by~\eqref{eq:fusion}
  \IF{$t = 0$} \STATE add $\boldsymbol{\eta}$ of~\eqref{eq:vis_innov} to the sums of~\eqref{eq:moments} \ENDIF
  \STATE $\bxhat^{t+1} \leftarrow \bxhat^t \bplus \delta\bx$ of~\eqref{eq:inekf_solve}
\ENDFOR
\STATE $\Phat \leftarrow (\bI - \bG)\Phat$;\;
       $\mathbf{c}_j \leftarrow \mathbf{A}\mathbf{c}_j - \sigma^2_{n,j}\mathbf{k}_j$;\;
       update the map and $\mathcal{W}$
\STATE at the end of an epoch, $\hat{\boldsymbol{\theta}} \leftarrow$
       solution of~\eqref{eq:moments}, $\boldsymbol{\theta}\geq\mathbf{0}$
\end{algorithmic}
\end{algorithm}

\subsection{Unified Joint InEKF Update}
\label{sec:update}

\begin{figure}[!t]
  \centering
  \includegraphics[width=\columnwidth]{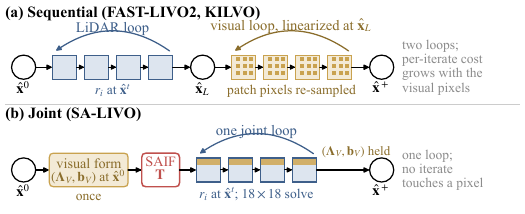}
  \caption{Update structure. (a)~Sequential designs iterate a LiDAR update,
    then a visual update linearized at its result $\hat{\mathbf{x}}_L$ that
    re-samples the patch pixels. (b)~SA-LIVO compresses vision once at the
    prior into $(\bLam_V,\bb_V)$ and runs one loop.}
  \label{fig:update_loop}
\end{figure}

The fused form drives one InEKF update in the full $18$-dimensional state.
With the propagated state $\bxhat^0$ and the $t$-th iterate $\bxhat^t$,
\begin{align}
  \mathbf{K} &\trigeq \bLam_f + \Phat^{-1},\qquad
  \bG = \mathbf{K}^{-1}\bLam_f,
  \label{eq:Kinv} \\
  \delta\bx &= \mathbf{K}^{-1}\bb_f
             + (\bxhat^0 \bminus \bxhat^t)
             - \bG\,(\bxhat^0 \bminus \bxhat^t),
  \label{eq:inekf_solve}
\end{align}
where $\mathbf{K}$ is the posterior information and $\mathbf{K}^{-1}$ the
posterior covariance; the first term is the measurement update, the second
anchors the iteration to the prior, and the third removes the part of that
offset that the measurement already explains (supplementary Section~S-I-B).
Unlike sequential designs (Fig.~\ref{fig:update_loop}), both sensors enter
one solve, vision is not linearized at a LiDAR-updated state, and one loop
of at most $K_{\max}=5$ iterations serves both.
An iteration costs one $18\times18$ solve and the re-evaluation of the
cached point-to-plane distances, whatever the number of visual
observations; $\mathbf{T}$ costs one eigendecomposition per frame.

The loop stops when $\|\delta\boldsymbol{\phi}\| < \epsilon_\phi$ and
$\|\delta\bp\| < \epsilon_p$ ($\epsilon_p = 0.15$~mm) hold for two
consecutive iterations, and the
covariance is updated as $\Phat \leftarrow (\bI_{18} - \bG)\Phat$ without
inflation.
Algorithm~\ref{alg:inekf} summarizes the update.

\section{Experiments}
\label{sec:experiments}

\subsection{Experimental Setup}
\label{sec:setup}

Our system is implemented in C++ with Robot Operating System (ROS) and evaluated on a laptop
(Intel i9-13900HX, 32~GB RAM) and an ARM platform (NVIDIA Jetson Orin,
12-core ARM Cortex-A78AE @ 2.2~GHz, 64~GB RAM).

\begin{figure}[!htbp]
  \centering
  \includegraphics[width=\columnwidth]{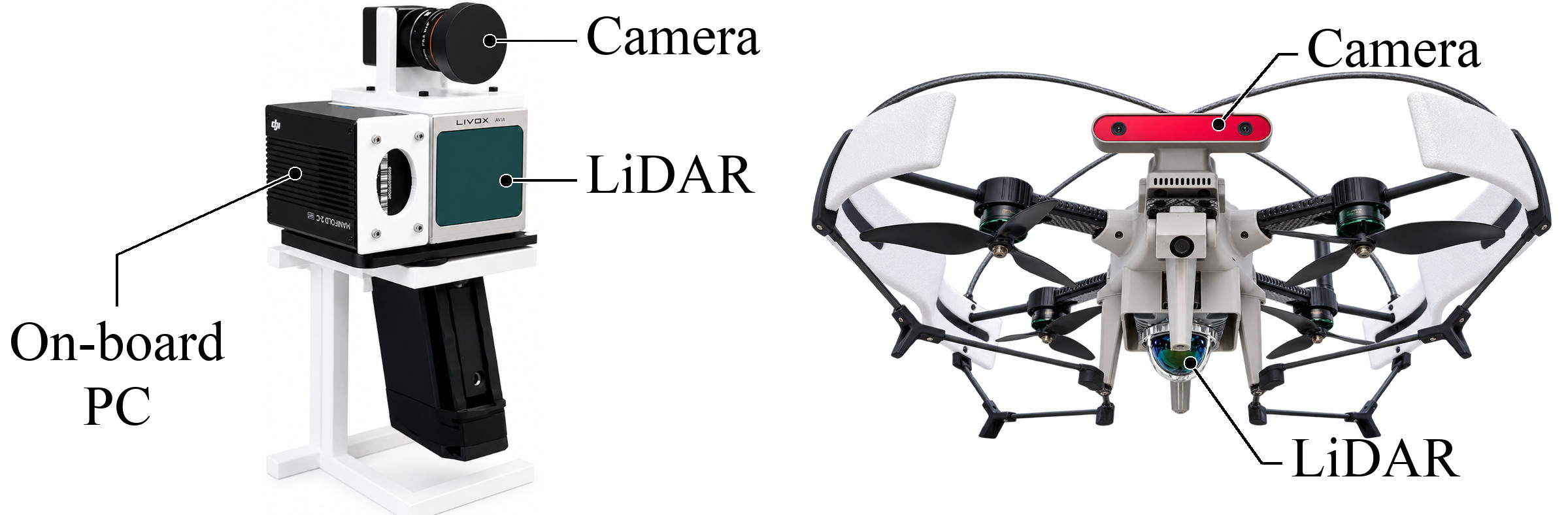}
  \caption{Data-collection platforms of the self-collected sequences:
    handheld (left) and UAV (right).}
  \label{fig:platform}
\end{figure}

Our self-collected sequences were acquired with
the two platforms of Fig.~\ref{fig:platform}.
The handheld platform carries a Livox AVIA LiDAR (10~Hz) with built-in IMU
(200~Hz) and an industrial camera (10~Hz), synchronized in hardware (PPS
time stamps, external camera trigger), and an on-board PC.
The UAV carries a Livox Mid-360 LiDAR (360\textdegree{}$\times$59\textdegree{},
10~Hz) with its built-in ICM40609 IMU (200~Hz) and an Intel RealSense D435i
camera (10~Hz, color images only) and flies 10--30~m above the ground.

We evaluate our system on 24 sequences of two public datasets.
HILTI~2022~\cite{zhang2022hilti} is recorded with a handheld sensor head
on a construction site and in historic buildings, with narrow stairwells,
cupolas, and long corridors; we use its 15 sequences other than the 80~GB
exp23, named after their scenes (official numbers in supplementary
Section~S-V-A).
MARS-LVIG~\cite{li2024marslvig} (9 sequences) is recorded from an unmanned
aerial vehicle at $80$--$90$~m above an airfield, an island, and a town.
Baselines are FAST-LIO2~\cite{xu2022fastlio2}, FAST-LIVO~\cite{zheng2022fastlivo},
FAST-LIVO2~\cite{zheng2022fastlivo2}, R3LIVE~\cite{lin2022r3live}, and
SR-LIVO~\cite{yuan2024srlivo}, using official open-source implementations.
For all methods, HILTI~2022 images are subsampled to the 10~Hz LiDAR rate,
the rate of MARS-LVIG, and on MARS-LVIG the runs start at cruise altitude
and end 2~s before the descent.
HILTI~2022 uses the official sparse-survey-point evaluator; MARS-LVIG uses
the translational absolute pose error (APE) against the real-time kinematic
(RTK) ground truth after SE(3) alignment and a search for the constant time
offset of each method.

\emph{Baseline reproduction.}
All baseline values in Table~\ref{tab:rmse} are our own runs, not values
copied from the original papers, so that every method is evaluated on the
same data with the same evaluator.
FAST-LIVO2 is run from the official repository
with the authors' HILTI~2022 configuration and calibration.
On Stairs the FAST-LIVO2 paper reports 0.016~m
\cite{zheng2022fastlivo2}; with the official configuration, and after
tuning its parameters for the sequence, we obtain 0.415~m.
The gap comes from one event, a match to another level of the stairwell
(Section~\ref{sec:accuracy}, Fig.~\ref{fig:exp03_drift}); we report our
reproduced value.
\begin{table}[!t]
  \centering
  \caption{Translational RMSE (m). $\times$: failed; green / yellow: best /
    second best; $^{\dagger}$: reproduced with the official code,
    \cite{zheng2022fastlivo2} reports 0.016 (Section~\ref{sec:setup}).}
  \label{tab:rmse}
  \scriptsize
  \renewcommand{\arraystretch}{0.9}
  \setlength{\tabcolsep}{2.4pt}
  \begin{tabular}{@{}c@{\ }l cccccc@{}}
    \toprule
    & & FAST- & FAST- & FAST- & & & \\
    & Seq. & LIO2 & LIVO & LIVO2 & R3LIVE & SR-LIVO & \textbf{Ours} \\
    \midrule
    \multirow{15}{*}{\rotatebox[origin=c]{90}{\textbf{HILTI'22}}}
     & Ground   & \second{0.014} & 0.019 & \best{0.011} & 0.084 & 0.015 & \best{0.011} \\
     & Multilevel     & 0.037 & \second{0.033} & \best{0.020} & 0.086 & 0.050 & \best{0.020} \\
     & Stairs         & \second{0.392} & 0.789 & 0.415$^{\dagger}$ & $\times$ & $\times$ & \best{0.029} \\
     & Upper-1    & 0.045 & 0.082 & \second{0.039} & 0.072 & \best{0.024} & \best{0.024} \\
     & Upper-2  & 0.044 & \second{0.010} & 0.012 & 0.080 & 0.012 & \best{0.008} \\
     & Upper-3  & 0.048 & 0.053 & \best{0.010} & 0.083 & \second{0.018} & \best{0.010} \\
     & Corridor               & 0.079 & 0.112 & \second{0.044} & 0.095 & \best{0.041} & 0.053 \\
     & Cupola-1                      & $\times$ & $\times$ & \second{0.167} & $\times$ & $\times$ & \best{0.108} \\
     & Cupola-2                    & $\times$ & $\times$ & \second{0.218} & $\times$ & $\times$ & \best{0.083} \\
     & Gallery-1               & 0.397 & 0.474 & \second{0.015} & $\times$ & 0.989 & \best{0.008} \\
     & Basement                  & 0.140 & 0.070 & \second{0.035} & 0.098 & $\times$ & \best{0.019} \\
     & Attic-1      & 2.227 & 0.741 & \second{0.192} & $\times$ & $\times$ & \best{0.094} \\
     & Attic-2    & 1.927 & 1.211 & \second{0.102} & $\times$ & $\times$ & \best{0.062} \\
     & Gallery-2    & 0.366 & $\times$ & \second{0.194} & $\times$ & $\times$ & \best{0.020} \\
     & Outside            & 0.121 & 0.076 & \second{0.036} & 0.088 & 0.048 & \best{0.025} \\
    \midrule
    \multirow{9}{*}{\rotatebox[origin=c]{90}{\textbf{MARS-LVIG}}}
     & AMtown01                            & 2.737 & 1.710 & \second{1.198} & $\times$ & 2.139 & \best{1.038} \\
     & AMtown02                            & 3.906 & 4.866 & \second{1.750} & $\times$ & 8.145 & \best{1.188} \\
     & AMtown03                            & 2.893 & 8.217 & \second{1.528} & $\times$ & $\times$ & \best{1.192} \\
     & HKairport01                         & $\times$ & $\times$ & \second{0.460} & $\times$ & $\times$ & \best{0.307} \\
     & HKairport02                         & $\times$ & $\times$ & \second{0.631} & $\times$ & $\times$ & \best{0.409} \\
     & HKairport03                         & $\times$ & $\times$ & \second{0.842} & $\times$ & $\times$ & \best{0.427} \\
     & HKisland01                          & 0.613 & 0.610 & \second{0.333} & $\times$ & 0.335 & \best{0.317} \\
     & HKisland02                          & $\times$ & $\times$ & \second{0.404} & $\times$ & $\times$ & \best{0.365} \\
     & HKisland03                          & $\times$ & $\times$ & \best{0.419} & $\times$ & $\times$ & \second{0.424} \\
    \bottomrule
  \end{tabular}
\end{table}

\subsection{Quantitative Accuracy}
\label{sec:accuracy}

Table~\ref{tab:rmse} reports the absolute translational root-mean-square
error (RMSE); ``$\times$''
marks a run that did not produce a complete trajectory, through process
termination or tracking loss, or that diverged (R3LIVE on MARS-LVIG: RMSE
above 100~m); such runs are excluded from all averages.

On HILTI'22, only our system and FAST-LIVO2 complete all 15 sequences; R3LIVE and
SR-LIVO each fail on seven of the indoor-degenerate runs.
Our system is the most accurate or tied on 14 of the 15 sequences; on the
long Corridor SR-LIVO is better by 1.2~cm.
The hardest are Stairs and the two cupola runs, where LiDAR and camera degrade together: near-coplanar scan
lines against featureless walls under low illumination.

\begin{figure}[!t]
  \centering
  \includegraphics[width=\columnwidth]{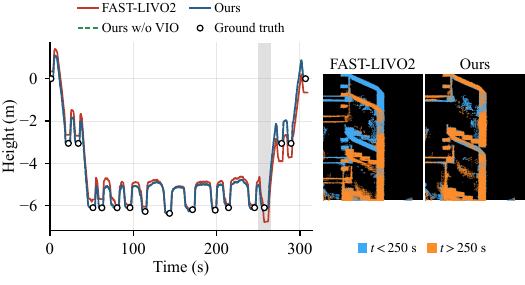}
  \caption{HILTI'22 Stairs. Left: height and ground-truth control points.
    Near 255~s (shaded) FAST-LIVO2 matches another level of the stairwell
    and stays 1~m low. Right: stairwell slice of the map from the raw scans;
    stairs mapped after the corner (orange) lie 1~m below those mapped before
    (blue) for FAST-LIVO2 and coincide for ours.}
  \label{fig:exp03_drift}
\end{figure}

On Stairs the result is decided by one event
(Fig.~\ref{fig:exp03_drift}).
Near 255~s the operator passes an unlit, narrow corner of the stairwell,
where the images are black and the scan geometry is degenerate.
There the FAST-LIVO2 estimate drops by about 1~m within a few seconds and
keeps this offset to the end: the sequence returns to its starting point,
and FAST-LIVO2 ends 1.0~m below it, whereas our system returns to within
1~cm of the starting height.
Our system without the visual term (w/o VIO) passes the corner in the same
way, so the difference comes from the LiDAR residual gate, not from the
visual module.
FAST-LIVO2 accepts a point-to-plane match within three standard deviations
of a variance that includes the state uncertainty, so its gate widens as the
uncertainty grows in the corner and admits matches to the planes of another
level of the stairwell; ours tests the residual against the plane and sensor variance
alone~\eqref{eq:sigmatot}, leaving the state uncertainty to the prior of the
joint update, and its gate does not widen.

On Cupola-2 our error is $2.6\times$ lower than that of FAST-LIVO2, the only
baseline to finish; on Attic-1 and -2 and Gallery-2 FAST-LIVO2 reaches
$0.10$--$0.19$~m and ours stays below 0.10~m.

MARS-LVIG stresses the opposite regime.
At $80$--$90$~m above flat terrain, the LiDAR constrains height and tilt
well but leaves the horizontal position weakly observed, and vision has to
carry it over long, featureless stretches.
Of the baselines, only FAST-LIVO2 completes all nine flights.
Our system is the most accurate on eight of the nine sequences and within
0.5~cm of FAST-LIVO2 on the ninth (HKisland03); on
the airfield it reduces the error of FAST-LIVO2 by $33$--$49\%$
(HKairport01--03).
The town flights (AMtown) are the longest, 5.4--7.4~km over flat farmland;
ours is the most accurate on all three, with an error of about 0.02\,\% of
the flight length.

\subsection{Ablation Study}
\label{sec:ablation}

Table~\ref{tab:ablation} removes or replaces one component at a time, in the
same binary and with each sequence's configuration, on the six degenerate
HILTI'22 sequences: SAIF replaced by the plain information sum; the
frame-level noise scale ($q\equiv1$); the Schmidt treatment of reused
templates replaced by the usual $1/n$ weighting of the $n$-th reuse; the
affine brightness model; the visual update altogether (LIO only); and the
joint update replaced by a LIO update followed by a VIO update.
On the nine other sequences all variants average 0.020--0.022~m; the
complete table, with two further variants, is given in supplementary
Section~S-V-A.

\begin{table}[!t]
  \centering
  \footnotesize
  \caption{Ablation on the degenerate HILTI'22 sequences: RMSE (m).}
  \label{tab:ablation}
  \renewcommand{\arraystretch}{0.95}
  \setlength{\tabcolsep}{2.2pt}
  \begin{tabular}{@{}l ccccccc@{}}
    \toprule
     & \textbf{Ours} & w/o & & reuse & w/o & LIO & LIO \\
    Seq. & \textbf{(full)} & SAIF & $q{\equiv}1$ & $1/n$ & affine & only & $\to$VIO \\
    \midrule
    Stairs & \textbf{0.029} & 0.137 & 0.151 & 0.047 & 0.133 & 0.052 & 0.131 \\
    Cupola-1 & \textbf{0.108} & 0.115 & 0.111 & 0.109 & 0.202 & 0.115 & 0.149 \\
    Cupola-2 & \textbf{0.083} & 0.135 & 0.112 & 0.122 & 0.118 & 0.085 & 0.095 \\
    Attic-1 & \textbf{0.094} & 0.108 & 0.100 & 0.126 & 0.170 & 0.114 & 0.208 \\
    Attic-2 & \textbf{0.062} & 0.077 & 0.085 & 0.145 & 0.076 & 0.085 & 0.092 \\
    Gallery-2 & \textbf{0.020} & 0.047 & 0.071 & 0.116 & 0.094 & 0.071 & 0.044 \\
    \midrule
    Mean  & \textbf{0.066} & 0.103 & 0.105 & 0.111 & 0.132 & 0.087 & 0.120 \\
    \bottomrule
  \end{tabular}
\end{table}

The full system is the most accurate on every degenerate sequence.
The most instructive comparison is with the LIO alone.
Admitting the visual information with its nominal confidence, i.e.\ without
SAIF or without the frame-level noise scale, gives a mean of 0.103 and
0.105~m, \emph{worse} than not using vision at all (0.087~m): an
over-confident visual term does more harm than a missing one.
With both error models the visual term becomes a net gain, lowering the
mean from 0.087 to 0.066~m.
On Stairs every other variant that uses vision, except the $1/n$ weighting,
ends between 0.13 and 0.15~m, against 0.029~m for the full system and 0.052~m for the LIO alone.
The Schmidt treatment matters most where observations stay in the window
longest relative to the motion: the $1/n$ weighting raises Attic-2 from 0.062
to 0.145~m and Gallery-2 from 0.020 to 0.116~m.
The sequential update is the worst variant on Attic-1 (0.208~m).
As expected for a direct method, removing the affine brightness model costs
most on the dark Cupola-1 (0.202~m).

\begin{figure*}[!htbp]
  \centering
  \includegraphics[width=\linewidth]{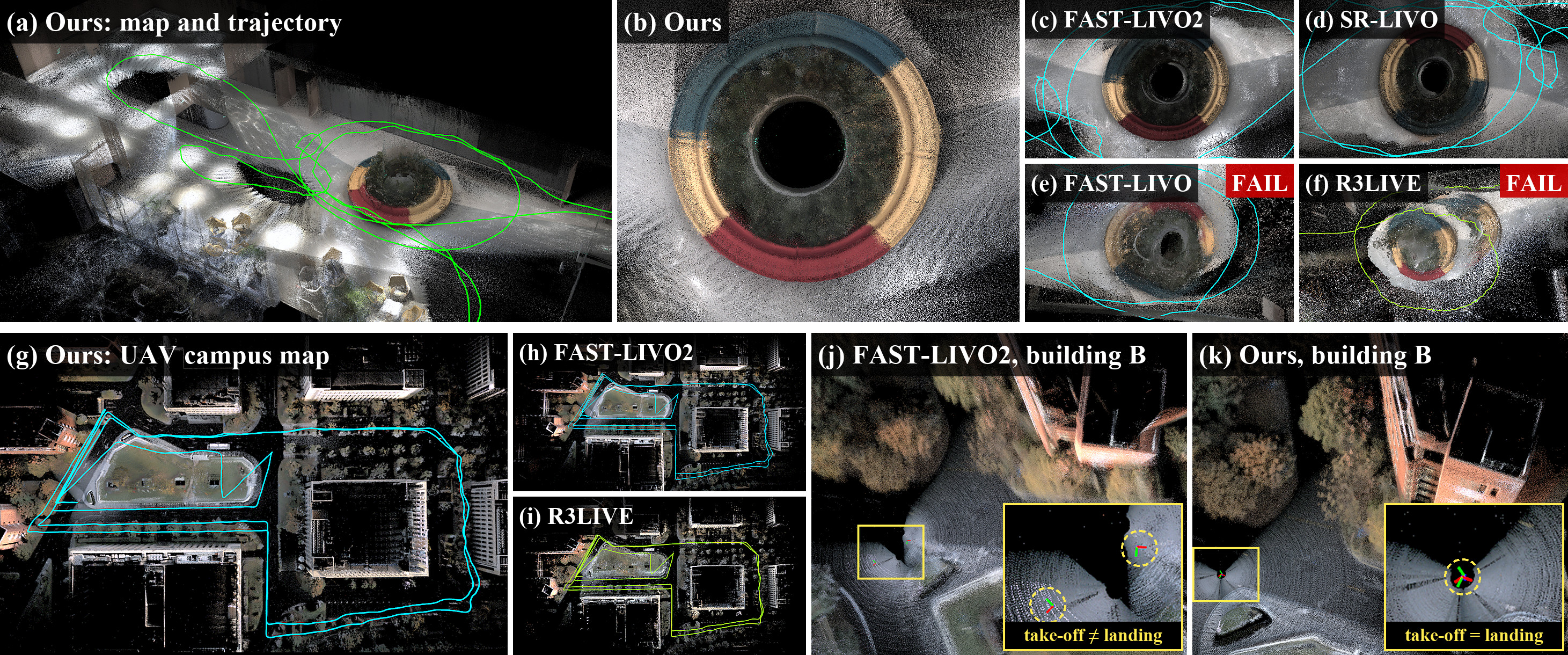}
  \caption{Maps on the self-collected sequences. Top, handheld
    \textit{indoor-chairs}: (a)~ours with trajectory; (b)--(f)~the central
    pillar by ours, FAST-LIVO2, SR-LIVO, FAST-LIVO, and R3LIVE (FAIL: doubled
    or smeared). Bottom, UAV \textit{campus}: (g)--(i)~global maps by ours,
    FAST-LIVO2, and R3LIVE; (j), (k)~building B and the take-off and landing
    poses (circled).}
  \label{fig:mapping}
\end{figure*}

\subsection{Mapping Quality on Self-Collected Sequences}

Fig.~\ref{fig:mapping} compares qualitative maps on our self-collected
sequences (platforms in Fig.~\ref{fig:platform}); supplementary Section~S-V-C
adds maps on MARS-LVIG, Oxford Spires, and a further UAV sequence.
On the handheld \textit{indoor-chairs} sequence, operating the sensor close to a central pillar yields near-degenerate LiDAR
geometry, with scan lines nearly co-planar and under-constrained along the
tangential direction.
FAST-LIVO doubles the pillar and R3LIVE smears it, and SR-LIVO and R3LIVE
do not complete the sequence.
In our system the visual information carries the tangential direction the
LiDAR leaves weak, admitted with the confidence its error model supports,
which keeps the pillar and its colored bands sharp.
On the UAV \textit{campus} sequence all three methods map the complete
building complex; R3LIVE's map is sparser and its trajectory visibly offset.
The difference lies in the sharpness of the structure surfaces.
When consecutive scans disagree in pose, the points of one surface spread
along its normal into a thick band; when they agree, the surface converges
to a thin shell.
FAST-LIVO2 tears and doubles building B, and its landing pose lies away
from the take-off pose although the UAV landed where it took off; in our
map building B remains intact and the two poses coincide.

\subsection{Runtime and Information Efficiency Analysis}
\label{sec:runtime}

\emph{Protocol.}
FAST-LIVO2, R3LIVE, and SR-LIVO are timed on the sequences they complete
(Table~\ref{tab:rmse}), with the shared parameters unified (no point
skipping; input downsampling $0.2$/$0.5$~m and map resolution $0.5$/$5.0$~m
on HILTI'22/MARS-LVIG; planes frozen at $100$ points in FAST-LIVO2 and
ours) and the others at the authors' values.
Sequences are replayed in real time, one method at a time, without map
sliding or visualization; memory is the resident set size at the end of the
sequence, and all timings are CPU-only.

\begin{table}[!t]
  \centering
  \caption{Per-frame time and final memory. $\times$: failed in
    Table~\ref{tab:rmse}; green / yellow: best / second best on the laptop.}
  \label{tab:perf_all}
  \scriptsize
  \renewcommand{\arraystretch}{0.9}
  \setlength{\tabcolsep}{1.5pt}
  \begin{tabular}{@{}c@{\ }l@{\ \ }cc@{\ \ }cc@{\ \ }cc@{\ \ }cc@{\ \ }cc@{}}
    \toprule
    & Seq. & \multicolumn{2}{c}{FAST-LIVO2} & \multicolumn{2}{c}{R3LIVE} & \multicolumn{2}{c}{SR-LIVO} & \multicolumn{2}{c}{\textbf{Ours}} & \multicolumn{2}{c}{Ours (ARM)} \\
    \cmidrule(lr){3-4} \cmidrule(lr){5-6} \cmidrule(lr){7-8} \cmidrule(lr){9-10} \cmidrule(lr){11-12}
             & & ms & GB & ms & GB & ms & GB & ms & GB & ms & GB \\
    \midrule
    \multirow{15}{*}{\rotatebox[origin=c]{90}{\textbf{HILTI'22}}}
     & Ground & 47.3 & \second{5.8} & \second{36.4} & 8.5 & 78.1 & 16.0 & \best{22.6} & \best{2.3} & 47.3 & 2.1 \\
     & Multilevel & 35.9 & \second{8.1} & \second{31.9} & 18.6 & 63.7 & 22.1 & \best{16.9} & \best{4.0} & 34.6 & 4.3 \\
     & Stairs & \second{17.1} & \second{2.8} & $\times$ & $\times$ & $\times$ & $\times$ & \best{9.8} & \best{0.9} & 22.4 & 1.0 \\
     & Upper-1 & 27.9 & \second{2.6} & \second{24.6} & 4.0 & 43.3 & 6.9 & \best{18.0} & \best{1.4} & 34.8 & 1.4 \\
     & Upper-2 & 26.1 & \second{2.2} & \second{24.8} & 3.8 & 44.2 & 6.4 & \best{17.6} & \best{1.2} & 42.1 & 1.3 \\
     & Upper-3 & \second{22.1} & \second{2.4} & 22.9 & 5.3 & 37.0 & 14.0 & \best{12.2} & \best{1.4} & 25.1 & 1.3 \\
     & Corridor & \second{14.2} & \second{1.0} & 16.4 & 3.9 & 31.0 & 1.9 & \best{11.2} & \best{0.8} & 22.7 & 0.8 \\
     & Cupola-1 & \second{10.7} & \second{3.9} & $\times$ & $\times$ & $\times$ & $\times$ & \best{6.9} & \best{1.3} & 14.7 & 1.2 \\
     & Cupola-2 & \second{14.5} & \second{3.2} & $\times$ & $\times$ & $\times$ & $\times$ & \best{10.8} & \best{1.2} & 20.5 & 1.2 \\
     & Gallery-1 & \second{23.9} & \second{1.8} & $\times$ & $\times$ & 46.7 & 3.7 & \best{15.2} & \best{1.0} & 30.7 & 1.0 \\
     & Basement & \second{9.4} & \second{0.6} & 17.6 & 2.5 & $\times$ & $\times$ & \best{6.0} & \best{0.4} & 12.3 & 0.4 \\
     & Attic-1 & \second{17.8} & \second{2.5} & $\times$ & $\times$ & $\times$ & $\times$ & \best{12.3} & \best{2.0} & 26.1 & 2.0 \\
     & Attic-2 & \second{18.0} & \second{2.0} & $\times$ & $\times$ & $\times$ & $\times$ & \best{12.1} & \best{1.5} & 28.8 & 1.4 \\
     & Gallery-2 & \second{13.4} & \second{1.2} & $\times$ & $\times$ & $\times$ & $\times$ & \best{7.8} & \best{0.7} & 14.3 & 0.7 \\
     & Outside & 34.5 & \second{2.7} & \second{30.9} & 6.2 & 67.0 & 9.0 & \best{21.6} & \best{1.2} & 46.4 & 1.2 \\
    \midrule
    \multirow{9}{*}{\rotatebox[origin=c]{90}{\textbf{MARS-LVIG}}}
     & AMtown01 & \second{61.3} & \second{10.9} & $\times$ & $\times$ & 80.0 & 21.4 & \best{24.7} & \best{5.5} & 44.8 & 5.3 \\
     & AMtown02 & \second{59.5} & \second{10.5} & $\times$ & $\times$ & 77.5 & 19.5 & \best{24.6} & \best{7.1} & 52.7 & 6.8 \\
     & AMtown03 & \second{63.6} & \second{10.9} & $\times$ & $\times$ & $\times$ & $\times$ & \best{25.7} & \best{7.7} & 47.9 & 8.1 \\
     & HKairport01 & \second{71.7} & \second{14.6} & $\times$ & $\times$ & $\times$ & $\times$ & \best{25.9} & \best{5.0} & 61.6 & 4.6 \\
     & HKairport02 & \second{73.1} & \second{8.9} & $\times$ & $\times$ & $\times$ & $\times$ & \best{26.6} & \best{5.3} & 49.5 & 5.8 \\
     & HKairport03 & \second{69.0} & \second{11.6} & $\times$ & $\times$ & $\times$ & $\times$ & \best{26.2} & \best{5.5} & 56.2 & 5.1 \\
     & HKisland01 & \second{30.8} & \second{2.9} & $\times$ & $\times$ & 55.7 & 10.0 & \best{15.6} & \best{0.9} & 32.4 & 0.9 \\
     & HKisland02 & \second{33.6} & \second{2.3} & $\times$ & $\times$ & $\times$ & $\times$ & \best{16.1} & \best{1.0} & 36.8 & 1.0 \\
     & HKisland03 & \second{31.6} & \second{2.1} & $\times$ & $\times$ & $\times$ & $\times$ & \best{15.9} & \best{1.1} & 29.7 & 1.1 \\
    \bottomrule
  \end{tabular}
\end{table}

On the seven HILTI'22 sequences that every method completes our system
averages 17.2~ms per frame, against 26.8~ms for R3LIVE, 29.7~ms for
FAST-LIVO2, and 52.0~ms for SR-LIVO, with $2.0\times$, $4.1\times$, and
$6.2\times$ less memory than FAST-LIVO2, R3LIVE, and SR-LIVO
(Table~\ref{tab:perf_all}).
On MARS-LVIG it is $2.0$--$2.8\times$ faster than FAST-LIVO2 with
$1.4$--$3.2\times$ less memory.
On the ARM board, without GPU, it needs $12.3$--$47.3$~ms per frame on
HILTI'22 and $29.7$--$61.6$~ms on MARS-LVIG, within the $100$~ms sensor
period on every sequence.

\emph{Time breakdown.}
The joint InEKF update takes about half of our per-frame time (6.5 of
13.4~ms on HILTI'22, 12.1 of 22.4~ms on MARS-LVIG), with the structure of
Fig.~\ref{fig:update_loop}(b).

\emph{Memory footprint.}
The LiDAR voxel maps of FAST-LIVO2 and ours have the same structure; the
difference is the visual map.
FAST-LIVO2 keeps every visual map point with up to $30$ patches, each
referring to a full image, whereas our observations store nine template
samples, only the last three images are kept, and the visual map is capped
at $10^4$ voxels.
R3LIVE and SR-LIVO additionally keep dense colored maps.

\section{Conclusion}
\label{sec:conclusion}

We presented SA-LIVO, a LiDAR-inertial-visual odometry system designed for
efficiency, whose fusion follows from explicit models of the visual errors
rather than hand-set thresholds.
SAIF models the error shared by all observations of a frame and admits the
visual information in each direction only up to the ceiling that error
implies, never attenuating the LiDAR; a frame-level noise scale and Schmidt
filtering of reused templates complete the visual error model.
One joint InEKF loop with cached visual Jacobians, frozen voxel planes, and
constant-time plane fitting bound the per-frame cost.

SA-LIVO is the most accurate or tied on 22 of the 24 sequences of HILTI'22
and MARS-LVIG and stays bounded where R3LIVE and SR-LIVO diverge; the
ablations show that admitting vision with its nominal confidence is worse
than not using it, and that the error models make it a net gain.
On the HILTI'22 sequences that every method completes it processes a frame
in 17.2~ms on a laptop processor, faster than all baselines with
$2.0$--$6.2\times$ less memory, and it runs within the 10~Hz sensor period on
an embedded ARM board.
As an odometry without loop closure, SA-LIVO accumulates drift with
distance, about 0.02\,\% of the flight length on the longest MARS-LVIG
flights.

\bibliographystyle{IEEEtran}

\begin{thebibliography}{99}

\bibitem{zheng2022fastlivo2}
C.~Zheng, W.~Xu, Z.~Zou, T.~Hua, C.~Yuan, D.~He, B.~Zhou, Z.~Liu, J.~Lin,
F.~Zhu, Y.~Ren, R.~Wang, F.~Meng, and F.~Zhang,
``FAST-LIVO2: Fast, direct LiDAR-inertial-visual odometry,''
\emph{IEEE Trans. Robot.}, vol.~41, pp.~326--346, 2025,
doi: 10.1109/TRO.2024.3502198.

\bibitem{lin2022r3live}
J.~Lin and F.~Zhang,
``R$^3$LIVE: A robust, real-time, RGB-colored, LiDAR-inertial-visual
tightly-coupled state estimation and mapping package,''
in \emph{Proc. IEEE Int. Conf. Robot. Autom. (ICRA)}, Philadelphia, PA,
USA, May 2022, pp.~10672--10678, doi: 10.1109/ICRA46639.2022.9811935.

\bibitem{shan2021lvisam}
T.~Shan, B.~Englot, C.~Ratti, and D.~Rus,
``LVI-SAM: Tightly-coupled lidar-visual-inertial odometry via smoothing
and mapping,''
in \emph{Proc. IEEE Int. Conf. Robot. Autom. (ICRA)}, Xi'an, China,
May 2021, pp.~5692--5698, doi: 10.1109/ICRA48506.2021.9561996.

\bibitem{zhang2016degeneracy}
J.~Zhang, M.~Kaess, and S.~Singh,
``On degeneracy of optimization-based state estimation problems,''
in \emph{Proc. IEEE Int. Conf. Robot. Autom. (ICRA)}, Stockholm, Sweden,
May 2016, pp.~809--816, doi: 10.1109/ICRA.2016.7487211.

\bibitem{hinduja2019degeneracy}
A.~Hinduja, B.-J.~Ho, and M.~Kaess,
``Degeneracy-aware factors with applications to underwater SLAM,''
in \emph{Proc. IEEE/RSJ Int. Conf. Intell. Robot. Syst. (IROS)},
Macau, China, Nov. 2019, pp.~1293--1299,
doi: 10.1109/IROS40897.2019.8968577.

\bibitem{zhang2022hilti}
L.~Zhang, M.~Helmberger, L.~F.~T. Fu, D.~Wisth, M.~Camurri,
D.~Scaramuzza, and M.~Fallon,
``Hilti-Oxford dataset: A millimetre accurate benchmark for simultaneous
localization and mapping,''
\emph{IEEE Robot. Autom. Lett.}, vol.~8, no.~1, pp.~408--415, Jan. 2023,
doi: 10.1109/LRA.2022.3226077.

\bibitem{li2024marslvig}
H.~Li, Y.~Zou, N.~Chen, J.~Lin, \emph{et al.}, ``{MARS-LVIG} dataset: A
multi-sensor aerial robots {SLAM} dataset for {LiDAR}-visual-inertial-{GNSS}
fusion,'' \emph{Int. J. Robot. Res.}, vol.~43, no.~8, pp.~1114--1127, 2024.

\bibitem{zhang2014loam}
J.~Zhang and S.~Singh,
``LOAM: Lidar odometry and mapping in real-time,''
in \emph{Proc. Robot.: Sci. Syst. (RSS)}, Berkeley, CA, USA, Jul. 2014,
doi: 10.15607/RSS.2014.X.007.

\bibitem{xu2022fastlio2}
W.~Xu, Y.~Cai, D.~He, J.~Lin, and F.~Zhang,
``FAST-LIO2: Fast direct LiDAR-inertial odometry,''
\emph{IEEE Trans. Robot.}, vol.~38, no.~4, pp.~2053--2073, Aug. 2022,
doi: 10.1109/TRO.2022.3141876.

\bibitem{zhang2025msclio}
T.~Zhang, M.~Yuan, L.~Wei, H.~Tang, and X.~Niu, ``{MSC-LIO}: An
{MSCKF}-based {LiDAR}-inertial odometry with same-plane cluster tracking,''
\emph{IEEE/ASME Trans. Mechatron.}, vol.~30, no.~6, pp.~7547--7557, 2025.

\bibitem{tuna2024xicp}
T.~Tuna, J.~Nubert, Y.~Nava, S.~Khattak, and M.~Hutter,
``X-ICP: Localizability-aware LiDAR registration for robust localization in
extreme environments,''
\emph{IEEE Trans. Robot.}, vol.~40, pp.~452--471, 2024,
doi: 10.1109/TRO.2023.3335691.

\bibitem{chen2026fastlivgo}
Z.~Chen, C.~Zheng, J.~Wen, X.~Zhang, J.~Xu, F.~Pan, and Y.~Cui,
``FAST-LIVGO: A degeneracy-robust LiDAR-inertial-visual-GNSS fusion
odometry,''
\emph{arXiv:2606.19190}, 2026.

\bibitem{mehra1972}
R.~K. Mehra, ``Approaches to adaptive filtering,'' \emph{IEEE Trans. Autom.
Control}, vol.~17, no.~5, pp.~693--698, 1972.

\bibitem{yuan2022voxelmap}
C.~Yuan, W.~Xu, X.~Liu, X.~Hong, and F.~Zhang,
``Efficient and probabilistic adaptive voxel mapping for accurate online
LiDAR odometry,''
\emph{IEEE Robot. Autom. Lett.}, vol.~7, no.~3, pp.~8518--8525, 2022,
doi: 10.1109/LRA.2022.3187250.

\bibitem{mourikis2007msckf}
A.~I.~Mourikis and S.~I.~Roumeliotis,
``A multi-state constraint Kalman filter for vision-aided inertial navigation,''
in \emph{Proc. IEEE Int. Conf. Robot. Autom. (ICRA)}, Roma, Italy, Apr. 2007,
pp.~3565--3572, doi: 10.1109/ROBOT.2007.364024.

\bibitem{bloesch2017rovio}
M.~Bloesch, M.~Burri, S.~Omari, M.~Hutter, and R.~Siegwart,
``Iterated extended Kalman filter based visual-inertial odometry using
direct photometric feedback,''
\emph{Int. J. Robot. Res.}, vol.~36, no.~10, pp.~1053--1072, 2017,
doi: 10.1177/0278364917728574.

\bibitem{qin2018vins}
T.~Qin, P.~Li, and S.~Shen,
``VINS-Mono: A robust and versatile monocular visual-inertial state
estimator,''
\emph{IEEE Trans. Robot.}, vol.~34, no.~4, pp.~1004--1020, Aug. 2018,
doi: 10.1109/TRO.2018.2853729.

\bibitem{engel2018dso}
J.~Engel, V.~Koltun, and D.~Cremers,
``Direct sparse odometry,''
\emph{IEEE Trans. Pattern Anal. Mach. Intell.}, vol.~40, no.~3,
pp.~611--625, 2018, doi: 10.1109/TPAMI.2017.2658577.

\bibitem{huang2010fej}
G.~P. Huang, A.~I. Mourikis, and S.~I. Roumeliotis,
``Observability-based rules for designing consistent EKF SLAM estimators,''
\emph{Int. J. Robot. Res.}, vol.~29, no.~5, pp.~502--528, 2010,
doi: 10.1177/0278364909353640.

\bibitem{hesch2014vio}
J.~A. Hesch, D.~G. Kottas, S.~L. Bowman, and S.~I. Roumeliotis,
``Consistency analysis and improvement of vision-aided inertial navigation,''
\emph{IEEE Trans. Robot.}, vol.~30, no.~1, pp.~158--176, Feb. 2014,
doi: 10.1109/TRO.2013.2277549.

\bibitem{song2023irvio}
Z.~Song, X.~Zhang, T.~Li, S.~Zhang, Y.~Wang, and J.~Yuan, ``{IR-VIO}:
Illumination-robust visual-inertial odometry based on adaptive weighting
algorithm with two-layer confidence maximization,'' \emph{IEEE/ASME Trans.
Mechatron.}, vol.~28, no.~4, pp.~1920--1929, 2023.

\bibitem{barrau2017invariant}
A.~Barrau and S.~Bonnabel,
``The invariant extended Kalman filter as a stable observer,''
\emph{IEEE Trans. Autom. Control}, vol.~62, no.~4, pp.~1797--1812, Apr. 2017,
doi: 10.1109/TAC.2016.2594085.

\bibitem{brossard2019consistency}
M.~Brossard, A.~Barrau, and S.~Bonnabel,
``Exploiting symmetries to design {EKF}s with consistency properties
for navigation and {SLAM},''
\emph{IEEE Sensors J.}, vol.~19, no.~4, pp.~1572--1579, Feb. 2019,
doi: 10.1109/JSEN.2018.2882714.

\bibitem{shi2023invlio}
P.~Shi, Z.~Zhu, S.~Sun, X.~Zhao, and M.~Tan,
``Invariant extended {K}alman filtering for tightly coupled
{LiDAR}-inertial odometry and mapping,''
\emph{IEEE/ASME Trans. Mechatron.}, vol.~28, no.~4, pp.~2213--2224, Aug. 2023,
doi: 10.1109/TMECH.2022.3233363.

\bibitem{tian2026eqf}
C.~Tian, F.~He, and N.~Hao, ``Equivariant filter transformations for
consistent and efficient visual--inertial navigation,'' \emph{IEEE/ASME
Trans. Mechatron.}, early access, 2026, doi: 10.1109/TMECH.2026.3730289.

\bibitem{zuo2019licfusion}
X.~Zuo, P.~Geneva, W.~Lee, Y.~Liu, and G.~Huang,
``LIC-Fusion: LiDAR-inertial-camera odometry,''
in \emph{Proc. IEEE/RSJ Int. Conf. Intell. Robot. Syst. (IROS)},
Macau, China, Nov. 2019, pp.~5848--5854,
doi: 10.1109/IROS40897.2019.8967746.

\bibitem{lin2021r2live}
J.~Lin, C.~Zheng, W.~Xu, and F.~Zhang,
``R$^2$LIVE: A robust, real-time, LiDAR-inertial-visual tightly-coupled
state estimator and mapping,''
\emph{IEEE Robot. Autom. Lett.}, vol.~6, no.~4, pp.~7469--7476, Oct. 2021,
doi: 10.1109/LRA.2021.3095515.

\bibitem{zheng2022fastlivo}
C.~Zheng, Q.~Zhu, W.~Xu, X.~Liu, Q.~Guo, and F.~Zhang,
``FAST-LIVO: Fast and tightly-coupled sparse-direct LiDAR-inertial-visual
odometry,''
in \emph{Proc. IEEE/RSJ Int. Conf. Intell. Robot. Syst. (IROS)},
Kyoto, Japan, Oct. 2022, pp.~4003--4009,
doi: 10.1109/IROS47612.2022.9981107.

\bibitem{gao2026kilvo}
J.~Gao, F.~Liu, T.~Zhang, and F.~Zha, ``{KILVO}: Kinematic--inertial--{LiDAR}--visual
odometry with robust multimodal adaptation for humanoid robots,''
\emph{IEEE/ASME Trans. Mechatron.}, early access, 2026,
doi: 10.1109/TMECH.2026.3721778.

\bibitem{lv2023ctlic}
J.~Lv, X.~Lang, J.~Xu, M.~Wang, Y.~Liu, and X.~Zuo, ``Continuous-time
fixed-lag smoothing for {LiDAR}-inertial-camera {SLAM},'' \emph{IEEE/ASME
Trans. Mechatron.}, vol.~28, no.~4, pp.~2259--2270, 2023.

\bibitem{wang2025givl}
X.~Wang, X.~Li, H.~Yu, H.~Chang, Y.~Zhou, and S.~Li, ``{GIVL-SLAM}: A robust
and high-precision {SLAM} system by tightly coupled {GNSS RTK}, inertial,
vision, and {LiDAR},'' \emph{IEEE/ASME Trans. Mechatron.}, vol.~30, no.~2,
pp.~1212--1223, 2025.

\bibitem{wen2026realvio}
X.~Wen, T.~Chao, J.~Yang, and M.~Yang, ``{REA-LVIO}: An asynchronous
{LiDAR}-visual-inertial odometry based on multistate constraint {K}alman
filter,'' \emph{IEEE/ASME Trans. Mechatron.}, early access, 2026,
doi: 10.1109/TMECH.2026.3704645.

\bibitem{schmidt1966}
S.~F. Schmidt, ``Application of state-space methods to navigation problems,''
in \emph{Advances in Control Systems}, vol.~3. New York: Academic Press, 1966,
pp.~293--340.

\bibitem{yuan2024srlivo}
Z.~Yuan, J.~Deng, R.~Ming, F.~Lang, and X.~Yang,
``SR-LIVO: LiDAR-inertial-visual odometry and mapping with sweep
reconstruction,''
\emph{IEEE Robot. Autom. Lett.}, vol.~9, no.~6, pp.~5110--5117, 2024,
doi: 10.1109/LRA.2024.3389415.

\end{thebibliography}

\clearpage
\includepdf[pages=-]{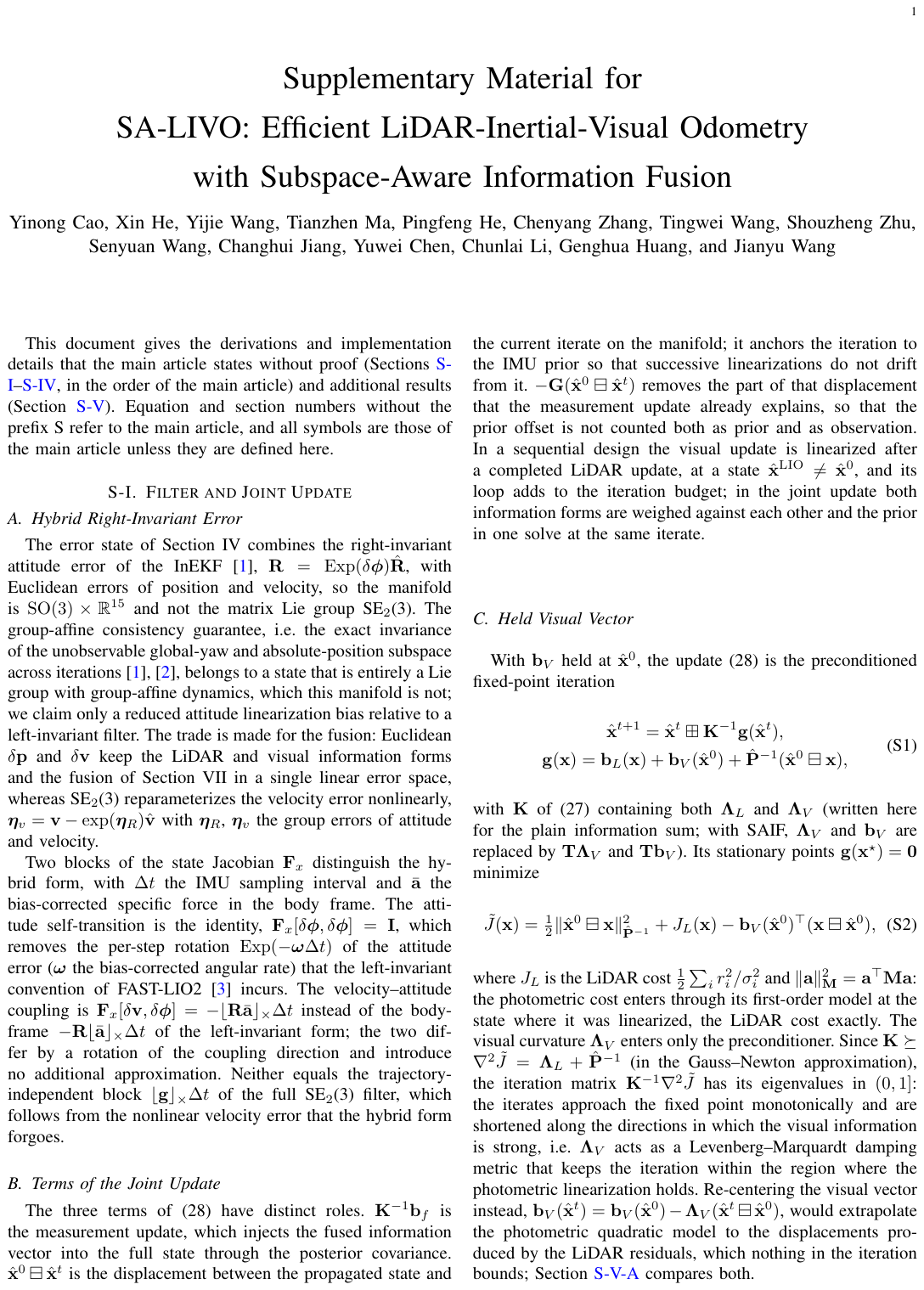}

\end{document}